\documentclass[runningheads]{llncs}

\usepackage{eccv}

\usepackage{eccvabbrv}

\usepackage{graphicx}
\usepackage{booktabs}

\usepackage{multirow}
\usepackage{array} 
\usepackage{tabularx}
\newcolumntype{Y}{>{\centering\arraybackslash}X}
\usepackage[utf8]{inputenc}
\usepackage{xcolor}
\usepackage{colortbl}
\usepackage{enumitem}
\usepackage{wrapfig}
\usepackage{lipsum} 
\usepackage{caption}
\usepackage{tikz}
\usepackage{pgfplots}
\pgfplotsset{compat=1.18}
\usepgfplotslibrary{groupplots}
\usepackage{amsmath}

\renewcommand{\topfraction}{0.9}
\renewcommand{\floatpagefraction}{0.8}
\usepackage{tcolorbox}

\definecolor{lightyellow}{RGB}{255, 232, 193}   
\definecolor{lightorange}{RGB}{255, 213, 193}   
\newcommand{\cellsecond}[1]{\begingroup\setlength{\fboxsep}{1pt}\colorbox{lightyellow}{\strut #1}\endgroup}
\newcommand{\cellbest}[1]{\begingroup\setlength{\fboxsep}{1pt}\colorbox{lightorange}{\strut #1}\endgroup}

\usepackage[accsupp]{axessibility}  

\usepackage[hidelinks]{hyperref}

\usepackage{orcidlink}

\begin{document}

\title{MorphGS: Morphology-Adaptive Articulated 3D Motion Transfer from Videos}

\titlerunning{MorphGS}

\author{Taeyeon Kim$^*$ \and
Youngju Na$^*$ \and
Jumin Lee \and
Sebin Lee \and
Minhyuk Sung \and
Sung-Eui Yoon$^\dagger$}

\authorrunning{T.~Kim et al.}

\institute{
KAIST, Republic of Korea\\
\email{\{tykim5931, yjna2907, jmlee, seb.lee, mhsung, sungeui\}@kaist.ac.kr}
}



\maketitle

\begingroup
\renewcommand{\thefootnote}{}
\footnotetext{$^*$Equal contribution \quad $^\dagger$Corresponding author}
\endgroup

\begin{figure}[!h]
    \centering
    \includegraphics[width=1.0\linewidth,trim=0 5.2cm 2cm 0,clip]{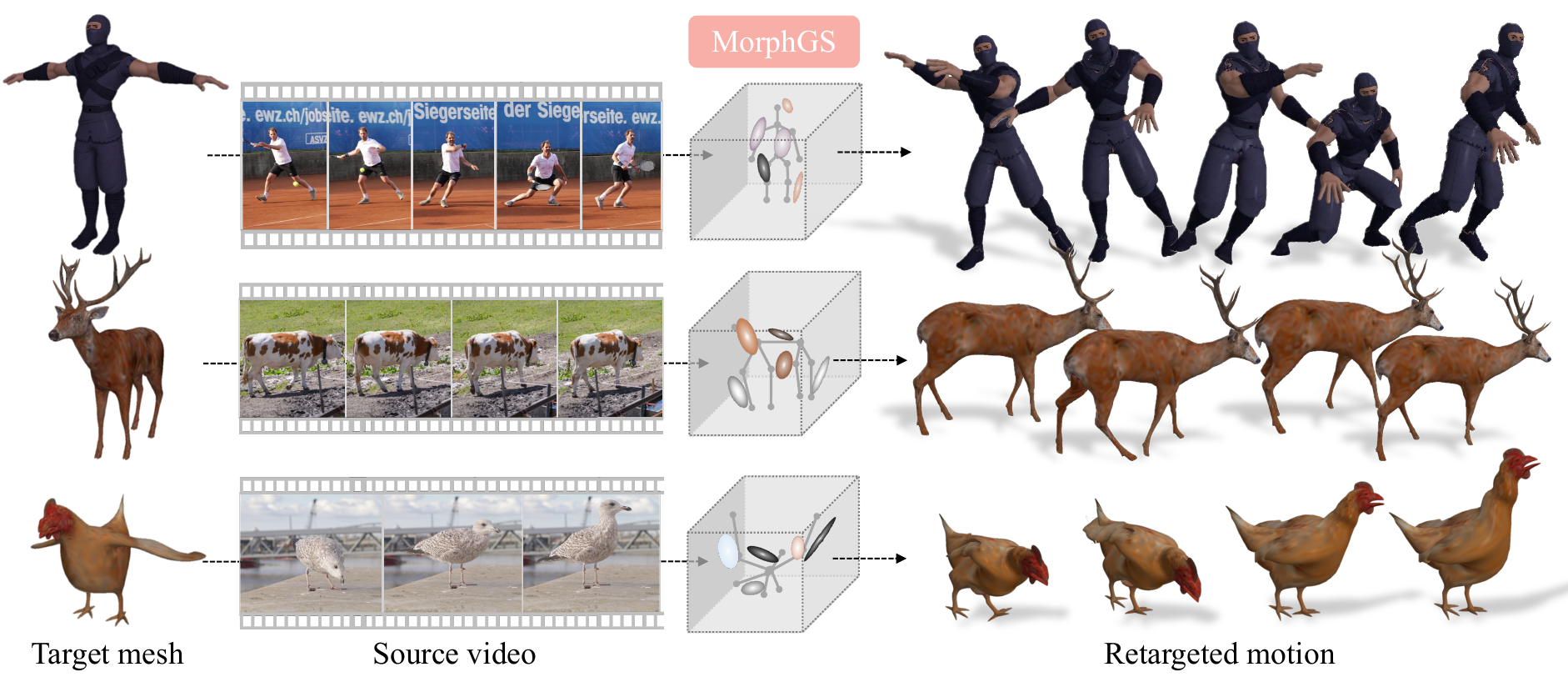}
    \caption{
    \footnotesize
    Given a monocular video and a rigged 3D target character, MorphGS directly optimizes target morphology and pose parameters to reproduce the observed motion without explicit source 3D reconstruction or category-specific parametric templates.}
    \label{fig:overview}
\end{figure}

\begin{abstract}
Transferring articulated motion from monocular videos to rigged 3D characters is challenging due to pose ambiguity in 2D observations and morphological differences between source and target. Existing approaches often follow a reconstruct-then-retarget paradigm, tying transfer quality to intermediate 3D reconstruction and limiting applicability to categories with parametric templates. 
We propose MorphGS, a framework that formulates motion retargeting as a target-driven analysis-by-synthesis problem, directly optimizing target morphology and pose through image-space supervision.
A rig-coupled morphology parameterization factorizes character identity from time-varying joint rotations, while dense 2D-3D correspondences and synthesized views provide complementary structural and multi-view guidance. Experiments on synthetic benchmarks and real-world videos show consistent improvements over baselines. Project page: \url{https://xodus777.github.io/MorphGS/}
\end{abstract}

\section{Introduction}
\label{sec:introduction}

Animating 3D characters from videos has long been a goal in vision and graphics. At its core, the task requires retargeting articulated motion observed in a monocular video to a rigged 3D target character. 
Despite progress in video-based 3D motion reconstruction, robust motion retargeting from monocular videos remains challenging to deploy broadly, as source videos and target characters often differ substantially in morphology and appearance~\cite{villegas2018nkn,aberman2020skeleton}.

A common paradigm follows a \emph{reconstruct-then-retarget}
pipeline~\cite{fu2024sync4d,maheshwari2023transfer4d,zhang2024magicpose4d,wang2023zero,muralikrishnan2024temporal}: first reconstruct an intermediate 3D representation from the source video, then transfer the recovered motion to a target character. 
These approaches often rely on parametric
templates~\cite{SMPL:2015,zuffi20173dsmal}, category-specific models~\cite{wu2023magicpony,jakab2024farm3d,yao2022lassie}, or
large-scale training data~\cite{li2024learning}, limiting their applicability to well-studied categories such as humans and quadrupeds.
Furthermore, the cascaded nature of the pipeline causes errors to propagate into the retargeting stage, where they are further amplified by morphological mismatches between the source and target.

We propose \textbf{MorphGS}, a video-to-3D motion transfer framework (Fig.~\ref{fig:overview}) that casts retargeting as a \emph{target-driven analysis-by-synthesis} problem~\cite{krull2015learning,beker2020monocular,wang2024neural}. 
Instead of reconstructing the source in 3D, we directly optimize the target character's morphology and motion so that its differentiable rendering~\cite{kerbl20233d} reproduces the source observations, while its structural identity remains anchored to the target.
This avoids both failure modes of the cascade, as no intermediate reconstruction propagates errors under morphological mismatch, and shape changes are grounded in the target rig rather than in category-specific priors.

The key to making this target-driven optimization tractable is to prevent pose updates from entangling with shape changes. MorphGS introduces an explicit morphology parameterization that factorizes time-invariant character identity from time-varying motion. Concretely, we model shape with structured morphology parameters---bone lengths, scale, and local rest-pose offsets anchored to the rig---while optimizing per-frame pose parameters via image-space supervision.
For robust structural guidance under monocular ambiguity, we combine two complementary signals. Dense 2D-3D semantic correspondences supply part-level anchors, while diffusion-based novel-view synthesis adds multi-view supervision.

Together, these components let MorphGS retarget motion without source reconstruction, motion priors, or category-specific training, while a rig-coupled parameterization keeps morphology and pose identifiable up to a global scale. 
We validate this design across synthetic and real-world benchmarks, showing that target-driven morphology-and-pose optimization provides an effective alternative to reconstruct-then-retarget pipelines.
\vspace{-5pt}
\section{Related Work}
\label{sec:related_works}

Our study addresses video-to-3D motion transfer, where the goal is to recover a pose sequence that animates a rigged 3D target mesh to reproduce articulated motion observed in videos. 
This task builds upon pose estimation from 2D observations and 3D motion transfer. 
We review each area below, including existing 2D-to-3D motion transfer, and position our approach within this landscape.

\paragraph{\textbf{Pose and shape estimation from videos}}
Recovering 3D pose from video often relies on category-specific parametric templates such as SMPL~\cite{SMPL:2015} for
humans~\cite{zhang2021pymaf, goel2023humans} and SMAL~\cite{zuffi20173dsmal} for quadrupeds~\cite{ruegg2023bite, lyu2024animer}. 
These methods achieve strong results but remain confined to categories with dense 3D annotations and parametric models.
Template-free methods~\cite{yao2022lassie, wu2023dove, wu2023magicpony, aygun2024saor, li2024learning} predict 3D shape and pose directly from image collections, yet their category-specific training limits generalization beyond seen categories and does not extend to cross-character retargeting.
Recent 4D reconstruction enables per-scene articulated modeling from monocular or sparse observations, leveraging representations such as 3D Gaussian splatting~\cite{kerbl20233d, Huang_2024_CVPR, Wu_2024_CVPR, Hu_2024_CVPR, yao2025riggs} and mesh-Gaussian hybrids~\cite{li2024dreammesh4d,wu2024sc4d}, while others animate static meshes via automatic rigging or video diffusion priors~\cite{song2025puppeteer,wang2025motiondreamer}.
However, these methods remain subject-specific, as the recovered motion is tied to the source identity and does not directly transfer to other characters.

\paragraph{\textbf{3D-to-3D motion transfer}}
While the above methods focus on recovering motion for a single subject, 
motion transfer addresses re-applying articulation
across different characters. 
Skeletal motion transfer methods operate on explicit skeleton representations~\cite{gleicher1998retargetting, villegas2018nkn,
aberman2020skeleton, villegas2021contact, chen2023weakly},
yielding robust results when two characters share compatible
rigs. 
Skeleton-free deformation methods~\cite{gao2018automatic, wang2020neural,
liao2022skeleton, wang2023zero, muralikrishnan2024temporal, yoo2024neural} relax the need for structural alignment, 
but both lines still require high-quality 3D motion sequences (\eg, skeletal trajectories or mesh animations), 
which remain scarce beyond well-studied categories such as humans.

\paragraph{\textbf{2D-to-3D motion transfer}}
Video-driven 3D motion transfer combines pose estimation and retargeting into a single pipeline. 
In the human and quadruped domains, parametric templates~\cite{SMPL:2015, zuffi20173dsmal} are paired with pose estimators~\cite{zhang2021pymaf, rueegg2022barc} and 3D transfer techniques~\cite{baran2007automatic,wang2023zero, muralikrishnan2024temporal}, but remain inherently limited to categories that such templates support.
Meanwhile, category-agnostic methods lift this restriction by first reconstructing source 3D motion with a generic prior---RGB-D videos~\cite{maheshwari2023transfer4d}, novel view synthesis~\cite{fu2024sync4d}, keypoint annotation~\cite{muralikrishnan2025smf}, or image-to-3D generative models~\cite{zhang2024magicpose4d, gong2026mocapanything}---and then transferring it to the target via algorithmic rigging or learned deformation.
Despite their generality, these approaches follow a \emph{reconstruct-then-retarget} paradigm whose performance is tied to the intermediate source reconstruction.
While recent works~\cite{muralikrishnan2025smf, gong2026mocapanything} reduce the need for dense reconstruction by learning deformation transfer from sparse 3D keypoints or skeletal joints, their generalization still depends on the shapes and motions seen during training.

In contrast, our method casts video-to-3D retargeting as a target-driven analysis-by-synthesis problem, 
directly optimizing the target's morphology and pose through image-space supervision via differentiable Gaussian rendering. 
A skeletal-anchored morphology parameterization accounts for the shape differences 
between source and target while preserving the target's rig, enabling 
transfer between morphologically different characters without parametric templates, motion priors, or explicit 3D supervision.
\section{Methods}
\label{sec:Methods}

\begin{figure*}[t]
    \centering
    \includegraphics[width=1.0\linewidth,trim={0 6.1cm 0cm 0},clip]{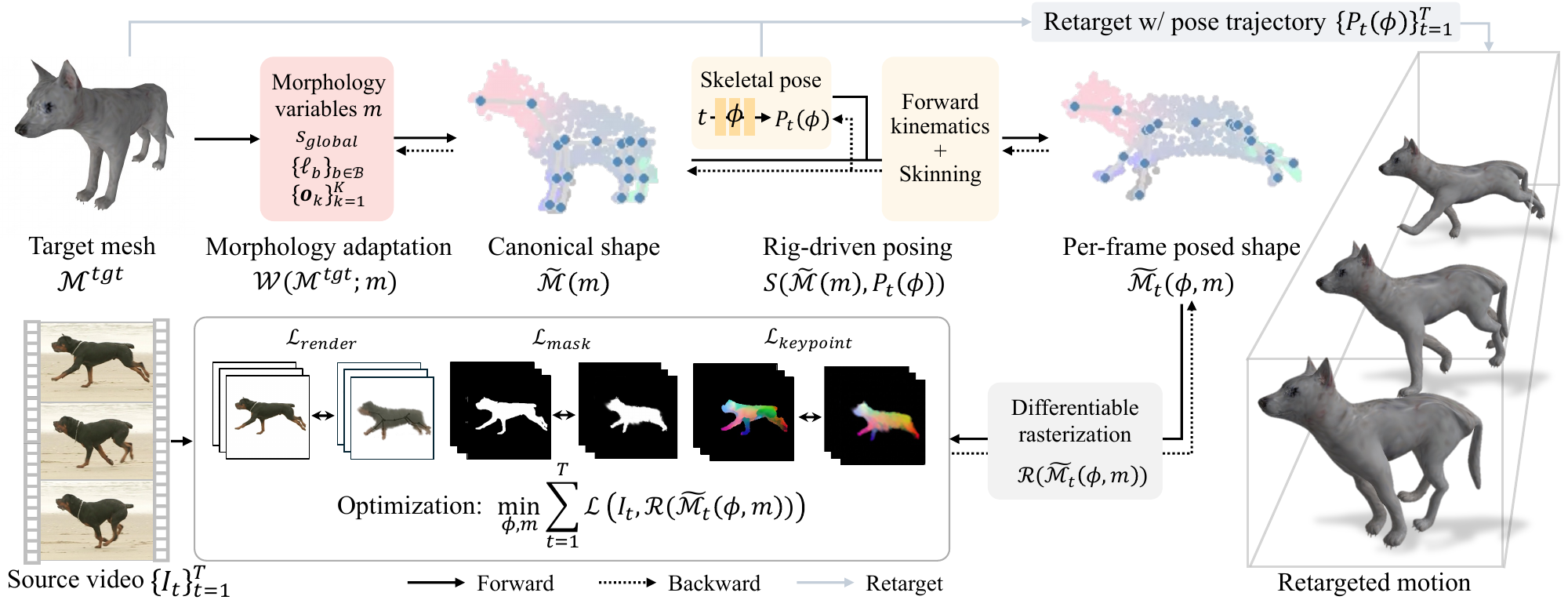}
    \caption{
    \footnotesize
    \textbf{Overview of MorphGS.} Given a source video and a target character, MorphGS first represents the target as a morphology-adaptive articulated 3D mesh (Sec.~\ref{sec:morph_param}). 3D Gaussian representation is wrapped to deform pose and render differentiably (Sec.~\ref{sec:rendering}). The morphology and pose parameters are jointly optimized through image-space supervision (Sec.~\ref{sec:optimization}). The recovered motion is then retargeted to the target character, yielding an animation that closely follows the source video.
    }
    \label{fig: pipeline}
\end{figure*}

\subsection{Problem Formulation}
\label{sec:method_overview}
Consider a source monocular RGB video $\{I_t \in \mathbb{R}^{H \times W \times 3}\}_{t=1}^{T}$ depicting an articulated subject.
Given a rigged target mesh $\mathcal{M}^{\mathrm{tgt}}$ whose morphology may differ from the source, our goal is to estimate a temporally coherent pose trajectory
$\{P_t\}_{t=1}^{T}$ that retargets the observed motion onto $\mathcal{M}^{\mathrm{tgt}}$ while preserving the target's structural identity  (i.e., skeleton topology and mesh connectivity).

\paragraph{\textbf{Target-driven analysis-by-synthesis}}
We formulate video-to-3D motion retargeting as a target-driven analysis-by-synthesis problem.
Instead of reconstructing the source in 3D and transferring motion afterward, we directly optimize
target-specific parameters so that the rendering of the animated target closely reproduces the input frames.
At each time step $t$, the posed shape $\tilde{\mathcal{M}}_t$ is obtained in two stages:
\textit{(i) morphology adaptation} in the target's canonical space, which adjusts limb proportions, global scale, and local geometry to bridge the shape gap between the source and target, followed by \textit{(ii) rig-driven posing} via forward kinematics and linear blend skinning:
\begin{equation}
\tilde{\mathcal{M}}(m)=\mathcal{W}(\mathcal{M}^{\mathrm{tgt}};m),\qquad
\tilde{\mathcal{M}}_t(\phi,m)=\mathcal{S}\!\big(\tilde{\mathcal{M}}(m),\, P_t(\phi)\big),
\label{eq:posed_state_factored}
\end{equation}
where $\mathcal{W}(\cdot\,;m)$ is the morphology adaptation
function parameterized by $m$, producing the adjusted canonical
shape $\tilde{\mathcal{M}}$, and $\mathcal{S}$ denotes the
skinning operator that applies skeletal pose $P_t(\phi)$ to
$\tilde{\mathcal{M}}$, yielding the per-frame posed shape
$\tilde{\mathcal{M}}_t$. Here $\phi$ denotes the parameters
of a temporally conditioned pose network.

A differentiable renderer $\mathcal{R}$ maps the posed shape $\tilde{\mathcal{M}}_t$ to a predicted frame $\hat{I}_t = \mathcal{R}(\tilde{\mathcal{M}}_t)$, where we adopt the 3D Gaussian rasterizer~\cite{kerbl20233d} for efficient rendering.
We estimate target-side parameters $(\phi,m)$ by minimizing objectives in the projection space over the foreground mask $\Omega_t$:
\begin{equation}
\min_{\phi,\, m}\ \sum_{t=1}^{T} \mathcal{L}\left(I_t,\ \mathcal{R}\big(\tilde{\mathcal{M}}_t(\phi,m)\big);\ \Omega_t\right).
\label{eq:target_objective_compact}
\end{equation}
Once optimization is complete, only the recovered pose trajectory $\{P_t(\phi)\}_{t=1}^T$ is applied back to the target rig $\mathcal{M}^{\mathrm{tgt}}$, preserving its original geometry. 
Note that we fix intrinsics and do not estimate per-frame extrinsics in our renderer. Global source–target alignment in each frame is approximated by a per-frame root transform in $SE(3)$.

Building on the above formulation, we first introduce a morphology-adaptive parameterization for articulated 3D modeling (Sec.~\ref{sec:morph_param}), then describe pose-conditioned Gaussian rendering (Sec.~\ref{sec:rendering}), and finally present the full objective and optimization procedure (Sec.~\ref{sec:optimization}). The overall pipeline is illustrated in Fig.~\ref{fig: pipeline}.

\subsection{Morphology-Parameterized Articulated 3D Modeling}
\label{sec:morph_param}
The goal of the morphology adaptation function $\mathcal{W}$ is to 
account for time-invariant shape differences between the source 
subject and the target character (\eg, differences in limb proportions, body size, or
local geometry), so that the per-frame source observations can
be explained primarily by the pose trajectory $\{P_t(\phi)\}_{t=1}^T$ through
the skinning operator $\mathcal{S}$
(Eq.~\eqref{eq:posed_state_factored}).

\begin{figure*}[t]
  \centering
  \includegraphics[width=\linewidth,trim={0 12.6cm 0cm 0},clip]{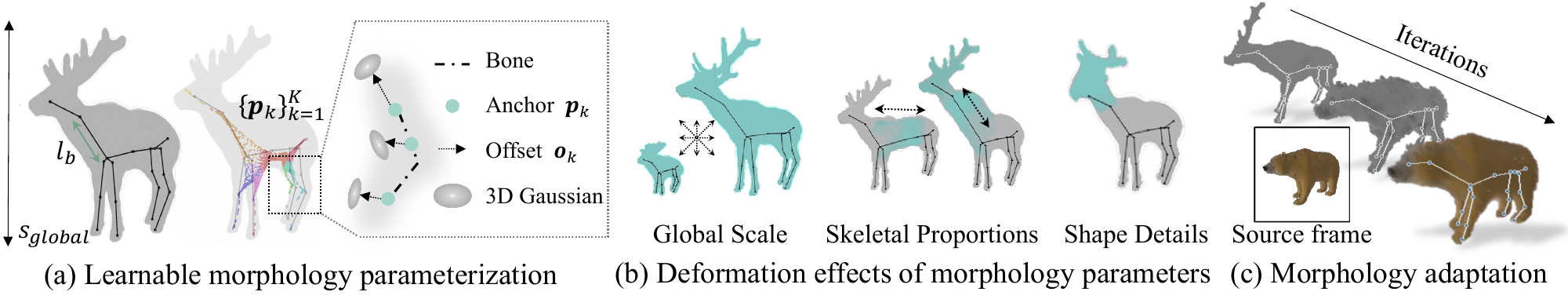}
    \caption{\textbf{Morphology parameterization.}
    (a) Skeleton anchor $\mathbf{p}_k$, local offset $\mathbf{o}_k$, and 3D Gaussian primitives on the target rig.
    (b) Distinct deformation induced by global scale, bone lengths, and offset Gaussians.
    (c) An example of morphology adaptation.}
  \label{fig:morphology}
\end{figure*}

However, jointly optimizing shape and pose from monocular observations is challenging due to the well-known shape-pose ambiguity, where shape changes can compensate for pose
updates in the image plane~\cite{xiao2004closed,hmrKanazawa17}.
If the canonical shape is optimized without structural and temporal constraints, it can absorb per-frame pose errors and lead to geometric drift.
We mitigate this by 
restricting all shape changes to structured \emph{time-invariant} morphology 
parameters coupled to the target rig, thereby reducing the degrees 
of freedom that can explain the same 2D observations. Accordingly, 
we parameterize morphology as:
\begin{equation}
m \;=\; \Big(s_{\mathrm{global}},\ \{\ell_b\}_{b\in\mathcal{B}},\ \{\mathbf{o}_k\}_{k=1}^{K}\Big),
\label{eq:morph_vars}
\end{equation}
where $s_{\mathrm{global}} \in \mathbb{R}_{+}$ resolves monocular depth--scale ambiguity, $\ell_b \in \mathbb{R}_{+}$ are learnable bone lengths controlling skeletal proportions, 
and $\mathbf{o}_k \in \mathbb{R}^{3}$ are local geometric offsets that capture residual shape details (\eg, limb thickness, silhouette) beyond what the skeleton alone can express, as illustrated in Fig.~\ref{fig:morphology}.
We assume the target mesh
$\mathcal{M}^{\mathrm{tgt}}$ is rigged with a fixed kinematic tree
and skinning weights $\{w_{kj}\}$. 
For unrigged targets (\eg, generated mesh), we obtain these using off-the-shelf auto-rigging methods~\cite{xu2020rignet}. 

\paragraph{\textbf{Learnable bone lengths}}
Let $\mathcal{J}$ denote the set of joints with root joint $j_{\mathrm{root}}$.
For each bone $b\in\mathcal{B}$, $\mathbf{d}_b$ denotes the unit direction from its parent to child joint.
We define the rest-pose joint locations $\mathbf{j}_{\mathrm{rest}}:\mathcal{J}\rightarrow\mathbb{R}^3$ as:
\begin{equation}
\mathbf{j}_{\mathrm{rest}}(j; \ell)
=
\mathbf{j}_{\mathrm{rest}}(j_{\mathrm{root}})
+
\sum_{b\in \mathcal{P}(j_{\mathrm{root}},\,j)} \ell_b\,\mathbf{d}_b,
\label{eq:rest_joints}
\end{equation}
where $\mathcal{P}(j_{\mathrm{root}},j)$ denotes the kinematic-chain path from the root to joint $j$.
The kinematic tree $\mathcal{B}$ and directions $\mathbf{d}_b$ are initialized from the target rig and kept fixed, while only the bone lengths $\ell_b$ are optimized.
This preserves the skeletal topology while allowing bone proportions to adapt via $\ell_b$ for all $b\in\mathcal{B}$.

\paragraph{\textbf{Canonical target shape}}
Given the adjusted skeleton, we define each canonical vertex 
position relative to the rig. Specifically, we compute a 
skeleton-dependent anchor $\mathbf{p}_k$ as the skinning-weighted 
average of the rest joints:
\begin{equation}
\mathbf{p}_k
=
\sum_{j\in\mathcal{J}} w_{kj}\,\mathbf{j}_{\mathrm{rest}}(j;\ell).
\label{eq:skeleton_anchor}
\end{equation}
With normalized skinning weights ($w_{kj}\!\ge\!0$, $\sum_j w_{kj}\!=\!1$), $\mathbf{p}_k$ is a convex combination of joint locations and therefore lies in the convex hull of $\{\mathbf{j}_{\mathrm{rest}}(j)\}$. 
We then model the canonical vertex as a residual around this anchor,
$\mathbf{v}_k = \mathbf{p}_k + \mathbf{o}_k$,
where $\mathbf{o}_k \in \mathbb{R}^{3}$ is a skeleton-anchored rest-space offset relative to $\mathbf{p}_k$ that captures local residual geometry (\eg, thickness, silhouette) beyond skeletal proportions.

Finally, we apply a uniform global scale
$s_{\mathrm{global}} \in \mathbb{R}_{+}$ to the canonical
shape, including both the skeleton and surface geometry, to
resolve the monocular depth-scale ambiguity:
\begin{equation}
\bar{\mathbf{v}}_k = s_{\mathrm{global}}\,\mathbf{v}_k.
\label{eq:global_scale}
\end{equation}

\paragraph{\textbf{Discussion}}
Theoretically, recovering both 3D shape and motion from monocular 2D trajectories with unconstrained non-rigid structure is ill-posed~\cite{xiao2004closed}. 
Under weak-perspective projection, our formulation instead imposes a piecewise-rigid kinematic model in which bone lengths $\{\ell_b\}$ and local rest-space offsets $\{o_k\}$ are strictly time-invariant~\cite{kovalenko2019structure}. 
This factorization confines temporal variation to articulated joint rotations, which discourages morphology parameters from compensating for pose errors and reduces shape--pose entanglement. 
With sufficiently non-degenerate motion, the resulting kinematic constraints make morphology and pose identifiable up to a global scale factor. While real videos often deviate from these ideal assumptions (e.g., perspective effects and soft-tissue deformation), the analysis provides a useful explanation for the empirical stability of our target-driven optimization. We include a proof sketch and further discussion in the appendix.

\subsection{Pose-conditioned Gaussian Rendering}
\label{sec:rendering}
Our target-driven formulation jointly optimizes both the morphology parameters $m$ and the pose parameters $\phi$ through image-space gradients.
We realize this by representing the morphology-adapted canonical shape as an
articulated 3D Gaussian set, posing it through forward kinematics
and Linear Blend Skinning (LBS), and rendering via 3D Gaussian
splatting~\cite{kerbl20233d}.

\paragraph{\textbf{Articulated 3D Gaussians}}
We represent the target mesh with $K$ 3D Gaussian primitives,
initialized from the mesh vertices in the rest pose so that each
Gaussian inherits its vertex's skinning weights
$\{w_{kj}\}_{j\in\mathcal{J}}$ directly from the rig. Each
primitive $k$ carries a canonical center
$\bar{\mathbf{v}}_k \in \mathbb{R}^3$
(Eq.~\ref{eq:global_scale}) together with the standard splatting
attributes: anisotropic scale, rotation vector, opacity, and
view-dependent color~\cite{kerbl20233d}. This initialization ensures that skeletal deformations and surface geometry remain tightly coupled, preventing
the Gaussians from drifting away from the underlying rig during
optimization.

\paragraph{\textbf{Pose-conditioned deformation}}
To encourage temporal coherence, we parameterize the pose
trajectory with a single time-conditioned network rather than
optimizing independent per-frame pose:
\begin{equation}
P_t(\phi)=\big(\{\boldsymbol{\theta}_j^t\}_{j\in\mathcal{J}},\,
\boldsymbol{\delta}_{\mathrm{root}}^t\big)
= f_{\phi}\!\left(\mathrm{emb}(t)\right),
\label{eq:pose_mlp}
\end{equation}
where $\mathrm{emb}(t)$ is a sinusoidal time
embedding~\cite{vaswani2017attention, mildenhall2021nerf} and
$f_{\phi}$ is a lightweight MLP that outputs per-joint relative
rotations $\boldsymbol{\theta}_j^t$ (axis-angle) and root
translation $\boldsymbol{\delta}_{\mathrm{root}}^t$. Forward
kinematics converts the local rotations into global joint
transforms
$\{\mathbf{T}_j^t \in \mathrm{SE}(3)\}_{j\in\mathcal{J}}$, which
then deform the canonical centers via LBS:
\begin{equation}
\mathbf{v}_k^t
=
\sum_{j\in\mathcal{J}}
w_{kj}\,\mathbf{T}_j^t\,\bar{\mathbf{v}}_k.
\label{eq:lbs_centers}
\end{equation}
\paragraph{\textbf{Rasterization}}
The posed Gaussians are projected onto the image plane and
composited via depth-sorted alpha
blending~\cite{kerbl20233d}, producing a rendered frame
$\hat{I}_t = \mathcal{R}(\tilde{\mathcal{M}_t}(\phi,m)\big) )$. This closes the differentiable
chain from $(m,\phi)$ to the pixel-level objective in
Eq.~\ref{eq:target_objective_compact}, enabling joint
morphology--pose optimization through image-space supervision
alone.

\subsection{Differentiable Optimization Framework}
\label{sec:optimization}
During optimization, the Gaussian representation serves as a 
differentiable proxy, enabling image-space gradients to flow back 
to both shape and pose through the rendering pipeline.
Unlike prior pipelines that assume aligned
source--target~\cite{yao2025riggs,Hu_2024_CVPR,AnimalAvatars2024}
or require canonical pose initialization~\cite{song2025puppeteer}, 
our approach jointly optimizes morphology and pose from a single monocular video without such assumptions. 
Our objective combines rendering losses, keypoint loss, and regularization:
\begin{equation}
  \mathcal{L}_t
  \;=\;
  \underbrace{\mathcal{L}_{\mathrm{rgb}}
  \;+\;
  \mathcal{L}_{\mathrm{mask}}}_{\mathcal{L}_{\mathrm{render}}}
  \;+\;
  \mathcal{L}_{\mathrm{keyp}}
  \;+\;
  \mathcal{L}_{\mathrm{mv}}
  \;+\;
  \mathcal{L}_{\mathrm{reg}}.
  \label{eq:per_frame}
\end{equation}

\paragraph{\textbf{Rendering losses}}
The rendering terms provide the primary gradient signal for motion and
appearance alignment by directly comparing the rendered target
$\hat{I}_t$ with the source frame $I_t$ in the image plane.
Specifically, $\mathcal{L}_{\mathrm{rgb}}$ enforces photometric
consistency within the foreground region, and
$\mathcal{L}_{\mathrm{mask}}$ aligns the rendered alpha mask
$\hat{\Omega}_t$ with a foreground segmentation $\Omega_t$ to
stabilize global outline.

\begin{figure*}[t]
  \centering
  \includegraphics[width=\linewidth,trim={0 9.1cm 0cm 0},clip]{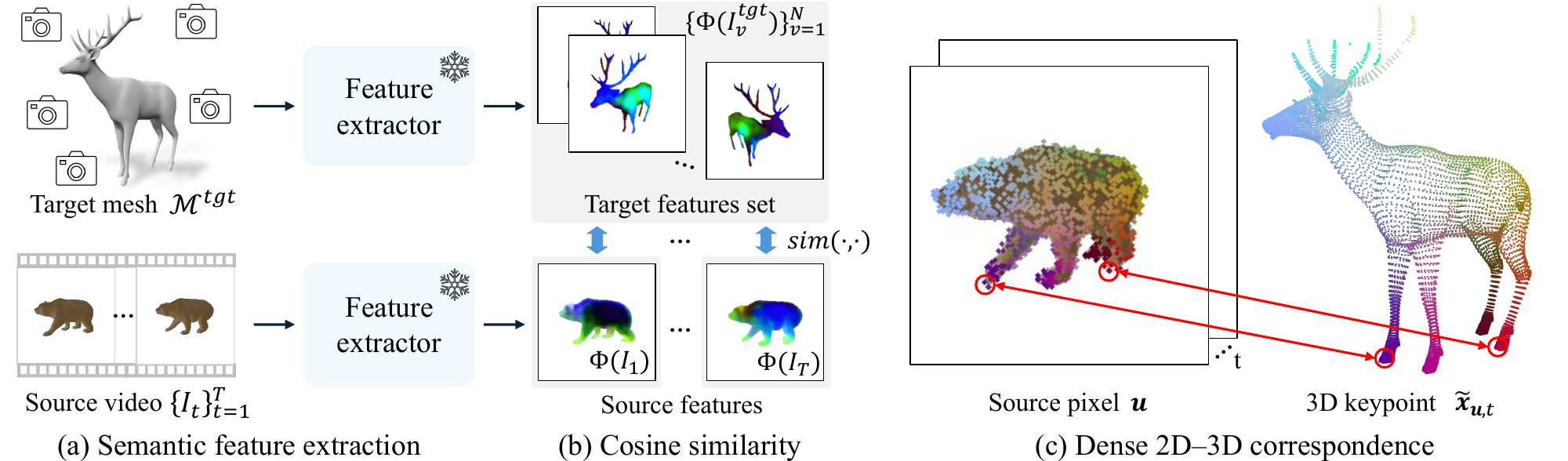}
 \caption{\textbf{Dense 2D--3D correspondence matching.} (a) Extracting semantic features from source frames and rendered target views. (b) Calculating similarity between them. (c) 3D keypoint mapping corresponding to the source pixel.}
  \label{fig:keypoint_loss}
\end{figure*}

\paragraph{\textbf{Keypoint loss}}

Photometric and mask supervision provide robust appearance and 
shape signals, but can be insufficient early in optimization or 
under large source--target morphological differences, where pixel-level losses alone struggle to establish reliable part-to-part correspondences.
To provide explicit structural guidance, we establish dense 2D--3D correspondences between the
source frame and the target mesh and use them as keypoint
supervision.

\noindent\emph{Dense 2D--3D correspondence.}
The detailed pipeline of our dense correspondence extraction module is illustrated in Fig.~\ref{fig:keypoint_loss}. We first extract dense semantic features from the source video frames $\{I_t\}_{t=1}^T$ and from the rendered images of the target mesh across $N$ views, $\{I_v^{\mathrm{tgt}}\}_{v=1}^N$, using a geometry-aware feature extractor $\Phi(\cdot)$~\cite{zhang2024telling}. 
For each frame $t$, we define $\mathbf{u}\in\mathbb{R}^2$ as a foreground pixel in the source frame $I_t$, and $\mathbf{x}\in\mathbb{R}^3$ as a vertex from the target mesh vertices $\mathcal{V}(\mathcal{M}^{\mathrm{tgt}})$.
Following SHIC~\cite{shtedritski2024SHIC}, we compute the pooled similarity score between $\mathbf{u}$ and $\mathbf{x}$:
\begin{equation}
\Sigma_t(\mathbf{u}, \mathbf{x}) = \operatorname*{pool}_{\substack{v, \ \mathbf{x} \in \mathcal{V}(\mathcal{M}^{\mathrm{tgt}})}} \, \mathrm{sim}\left( \Phi(I_t)[\mathbf{u}], \, \Phi(I_v^{\mathrm{tgt}})\left[\pi_v(\mathbf{x})\right] \right),
\label{eq:shic_pool}
\end{equation}
where $\mathrm{sim}(\cdot,\cdot)$ denotes cosine similarity,
$\pi_v(\mathbf{x})$ projects $\mathbf{x}$ onto rendered
view $I_v^{\mathrm{tgt}}$, and $\operatorname*{pool}$ operator aggregates scores
across all $N$ rendered views. Each source
pixel is then assigned to its best-matching 3D keypoint $\tilde{\mathbf{x}}_{\mathbf{u},t}$:
\begin{equation}
\tilde{\mathbf{x}}_{\mathbf{u},t}
=
\arg\max_{\mathbf{x} \in
\mathcal{V}(\mathcal{M}^{\mathrm{tgt}})}
\Sigma_t(\mathbf{u}, \mathbf{x}).
\label{eq:shic_max}
\end{equation}

\noindent\emph{Keypoint supervision.}
Let  $\mathcal{P}_t$ denote the set of sampled foreground pixels in frame $t$.
For each frame $t$, we select source pixels
$\mathbf{u}_i\in\mathcal{P}_t$, whose similarity scores exceed a confidence threshold, and penalize the
reprojection error between their observed locations and the
projected positions of the matched 3D keypoints:
\begin{equation}
  \mathcal{L}_{\mathrm{keyp}}
  =
  \frac{1}{\sum_t |\mathcal{P}_t|}
  \sum_{t}
  \sum_{\mathbf{u}_i\in\mathcal{P}_t}
  \rho\left(\mathbf{u}_i
  - \pi\bigl(\tilde{\mathbf{x}}_{\mathbf{u}_i,t}\bigr)\right),
  \label{eq:keypoint_loss}
\end{equation}
where $\pi(\cdot)$ denotes a fixed camera projection, and
$\rho(\cdot)$ is a smooth-$\ell_1$ penalty.

\paragraph{\textbf{Synthesized-view guidance}}
Monocular input leaves occluded body parts under-constrained, 
making pose ambiguous under self-occlusion. To provide 
complementary viewpoints, we leverage a diffusion-based novel-view 
synthesis module~\cite{yao2025sv4d} to generate temporally 
consistent pseudo-views of the source sequence. Our target-driven 
formulation naturally accommodates multi-view supervision, as the 
synthesized views simply provide additional image-space objectives 
over the same target parameters. We extend the rendering losses 
to all viewpoints $\mathcal{C}$:
\begin{equation}
  \mathcal{L}_{\mathrm{mv}}
  \;=\;
  \sum_{t=1}^{T}\sum_{c\in\mathcal{C}}
  \mathcal{L}_{\mathrm{render},t}^{(c)}.
  \label{eq:mv_loss}
\end{equation}
This additional supervision regularizes pose estimation under self-occlusion, where the monocular input alone provides insufficient constraints.

\vspace{-5pt}
\paragraph{\textbf{Regularizations}} Finally, $\mathcal{L}_{\mathrm{reg}}$ enforces frame-to-frame temporal motion smoothness. Additional details on regularizations are provided in the Appendix.
\section{Experiments}
\label{sec:experiments}

\subsection{Implementation details}
We employ a staged optimization strategy that first resolves global alignment (root translation, root orientation, $s_{\mathrm{global}}$), then we enable updating $\ell_b$ and local joint rotation estimation from 500 to 1.5K iterations. Lastly, from 1.5K iterations, we jointly optimize all parameters, including Gaussian splatting attributes (offsets $\mathbf{o}_k$, color, scale, rotation).
All experiments are run on a single NVIDIA RTX 4090 GPU. For a target mesh with $10$K vertices, the optimization process takes approximately 5 minutes for 5K iterations using Adam~\cite{kingma2014adam} with adaptive learning rates. Further details are provided in the Appendix.

\begin{table}[t]
    \centering
    \caption{\textbf{Quantitative evaluation on Mixamo and DT4D
    datasets.} Our method consistently outperforms all baselines
    across evaluated datasets on average. Results are averaged across scenes,
    with per-scene results in the appendix. PMD
    ($\times 10^{3}$, $\downarrow$) measures geometric accuracy
    and FVMD ($\times 10^{-3}$, $\downarrow$) measures
    perceptual motion fidelity.}
    \label{tab:quantitative}
    {\scriptsize
    \renewcommand{\arraystretch}{1.15}
    \setlength{\tabcolsep}{10pt}
    \newcommand{\con}{\textcolor{gray!70}{--}}
    \begin{tabular}{@{}l cc cc cc@{}}
        \toprule
        & \multicolumn{2}{c}{Mixamo-Humanoids} &
        \multicolumn{2}{c}{DT4D-Quadrupeds} &
        \multicolumn{2}{c}{DT4D-Others} \\
        \cmidrule(lr){2-3} \cmidrule(lr){4-5} \cmidrule(lr){6-7}
        Method & PMD & FVMD & PMD & FVMD & PMD & FVMD \\
        \midrule
        SPT$^+$       & 2.96 & 12.56 &
        \con & \con & \con & \con \\
        NPR$^+$       & 3.98 & 16.67 &
        3.28 & 16.32 & \con & \con \\
        Transfer4D    & 7.65 & 16.16 &
        4.67 & 15.34 & 7.16 & 17.52 \\
        Pinocchio$^+$ & 4.55 & 14.03 &
        5.56 & 15.24 & 7.59 & 14.23 \\
        \midrule
        Ours          & \textbf{1.91} & \textbf{8.82} &
        \textbf{1.33} & \textbf{8.81} &
        \textbf{1.95} & \textbf{12.83} \\
        \bottomrule
    \end{tabular}
    }
\end{table}

\subsection{Experimental Setup}
\paragraph{\textbf{Datasets}}
We evaluate MorphGS on both real-world and synthetic benchmarks to assess motion transfer performance across different categories.
For synthetic evaluation, we test on Mixamo~\cite{mixamo} and DeformingThings\nobreakdash-4D (DT4D)~\cite{dt4d}, following prior motion-transfer protocols~\cite{liao2022skeleton,yoo2024neural}.

In the Mixamo dataset, we utilize 12 humanoid source--target pairs with different motion types including diverse human motions (\eg, jumping, running, and dancing), where rigged target meshes are directly available from the dataset.
For DT4D, we include 20 animal source--target pairs with paired posed mesh sequences for the same motion, split into two groups by skeleton topology: \emph{DT4D-Quadrupeds} (four-legged animals), and \emph{DT4D-Others}, which includes 5 categories beyond the standard quadruped skeleton topology. Since DT4D does not provide rigged meshes, we obtain target rigs using off-the-shelf auto-rigging tools~\cite{xu2020rignet}.
To generate source videos, we animate the source mesh with the ground-truth pose sequence and render a monocular video from a fixed viewpoint using the differentiable mesh renderer in PyTorch3D~\cite{ravi2020pytorch3d}, yielding 1{,}505  source--target mesh pairs in total across the three splits.

For real-world evaluation, we curate 8 object-centric clips from DAVIS~\cite{davis}, comprising 4 human and 4 animal sequences. These clips span challenging capture conditions (e.g., motion blur, occlusion) and varying motion complexity (e.g., fast or non-rigid motion), as illustrated in Fig.~\ref{fig:qual_real}. We use the provided object masks when available and otherwise extract them with SAM3~\cite{sam3}, cropping each frame to the mask region. Extended real-world evaluations on 16 additional clips from SoloDance~\cite{solodance} and AiM~\cite{AiM} are provided in the appendix.

\paragraph{\textbf{Evaluation metrics}}
For synthetic datasets, we measure Point-wise Mesh Distance (PMD)~\cite{zhou2020unsupervised,wang2020neural} across all frames to evaluate motion transfer accuracy as the average per-vertex $\ell_2$ distance between the retargeted and ground-truth meshes following~\cite{liao2022skeleton,wang2023zero,yoo2024neural}.
For both synthetic and real datasets, we measure Fr\'echet Video Motion Distance (FVMD)~\cite{liu2024fr}, which measures perceptual motion plausibility by comparing distributions of velocity and acceleration features extracted from keypoint trajectories in rendered videos.

\paragraph{\textbf{Baselines}} We compare against baselines from two groups. (i) The first follows a \textit{reconstruct-then-retarget} paradigm, where a 3D mesh sequence is first reconstructed from the source RGB video using off-the-shelf parametric template pose estimators~\cite{goel2023humans,lyu2024animer}, and then retargeted to the target character via SPT~\cite{liao2022skeleton} (denoted SPT$^+$), NPR~\cite{yoo2024neural} (denoted NPR$^+$), or Pinocchio~\cite{baran2007automatic} (denoted Pinocchio$^+$).
(ii) The second is Transfer4D~\cite{maheshwari2023transfer4d}, which estimates a motion skeleton from RGB-D input and transfers it to the target via automatic rigging.

\begin{figure*}[t]
    \centering
    \includegraphics[width=\linewidth,trim={0 4.6cm 0cm 0},clip]{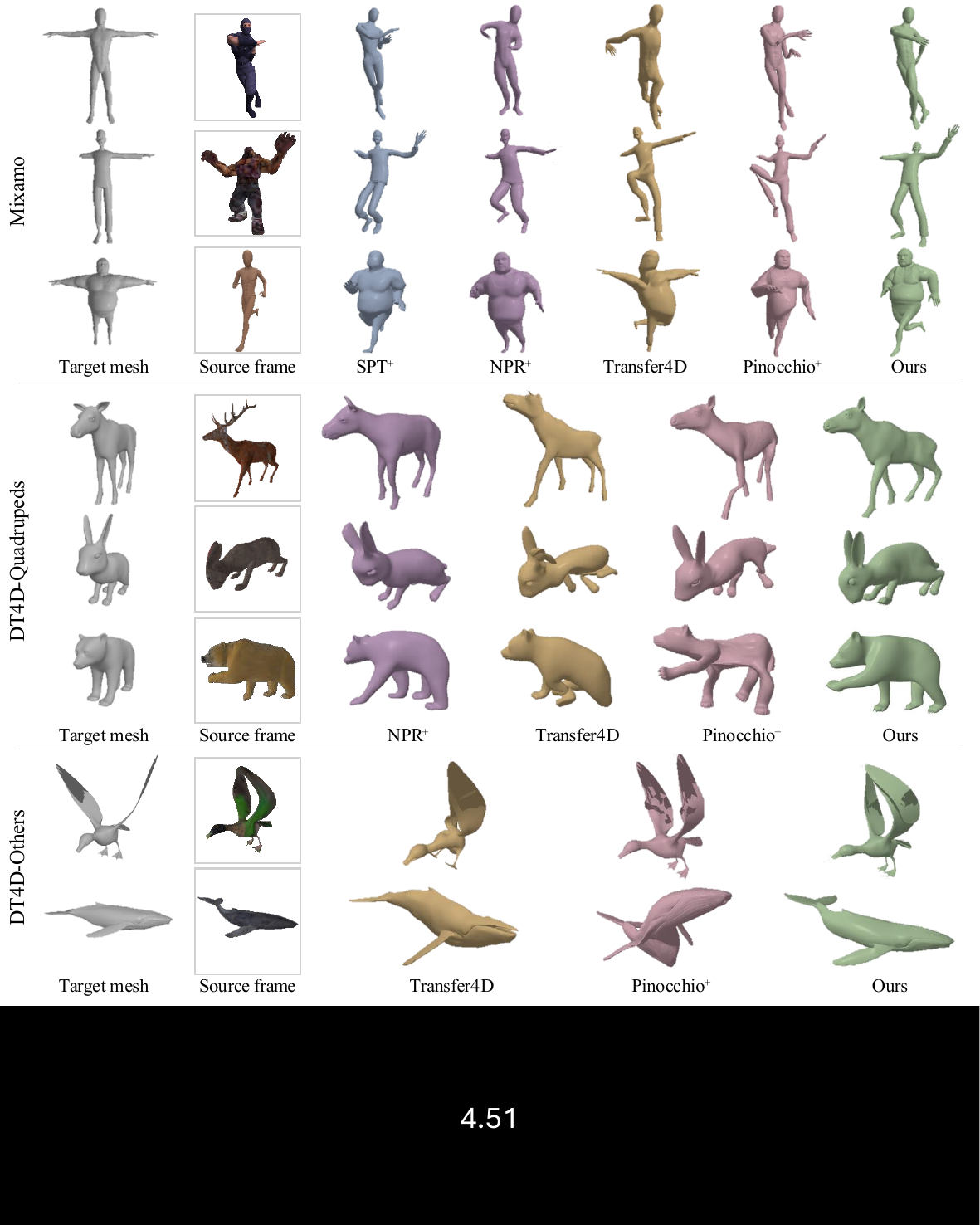}
    \caption{\footnotesize
    \textbf{Qualitative results on Mixamo and DT4D datasets.}
    Our method shows superior pose alignment compared to baselines across diverse objects.}
    \label{fig:qual_synthetic}
    \vspace{-15pt}
\end{figure*}

\subsection{Motion Transfer Evaluation}
\label{sec:experiments-quantitative}
We evaluate motion transfer on synthetic benchmarks spanning humanoid and animal targets with different topologies.
Tab.~\ref{tab:quantitative} summarizes the results.

\paragraph{\textbf{Humanoid motion transfer comparisons}}
For reconstruct-then-retarget baselines, we first estimate SMPL~\cite{SMPL:2015} pose sequences from the source video using an off-the-shelf monocular human pose estimator~\cite{goel2023humans}, and then retarget the recovered 3D motion to the target character using SPT~\cite{liao2022skeleton}, NPR~\cite{yoo2024neural}, and Pinocchio~\cite{baran2007automatic}, respectively.
For Transfer4D~\cite{maheshwari2023transfer4d}, which requires RGB-D input, we additionally provide rendered metric depth maps.

As shown in the first column of Tab.~\ref{tab:quantitative}, MorphGS shows lower PMD and FVMD than the compared baselines, consistent with the qualitative results in Fig.~\ref{fig:qual_synthetic}.
Baseline methods often exhibit incomplete motion transfer (\eg, third row of Fig.~\ref{fig:qual_synthetic}) or over-smoothed deformations. This is because they rely on learned pose priors or latent manifold constraints, and the retargeting stage may prioritize globally plausible motion over fine-grained vertex-level alignment.
In addition, preprocessing targets into manifold mesh representations~\cite{liao2022skeleton,yoo2024neural} can remove high-frequency surface details.
MorphGS instead optimizes target morphology and pose via differentiable rendering in the target parameter space, which maintains target mesh structure while achieving accurate pose alignment to source motions as in Fig.~\ref{fig:qual_synthetic}.

\begin{table*}[t]
  \centering
  \caption{\textbf{Quantitative evaluation on real-world videos.}
  We use FVMD ($\times 10^{-3}$, $\downarrow$) as
  the quantitative metric for motion fidelity.}
  \vspace{-5pt}
  \label{tab:quantitative-real}
  {\scriptsize
  \renewcommand{\arraystretch}{1.15}
  \setlength{\tabcolsep}{3.2pt}
  \newcommand{\con}{\textcolor{gray!70}{--}}
  \begin{tabular}{@{}l ccccc ccccc@{}}
    \toprule
    & \multicolumn{5}{c}{Human scenes} &
    \multicolumn{5}{c}{Animal scenes} \\
    \cmidrule(lr){2-6}\cmidrule(lr){7-11}
    Method & tennis & rollerblade & hockey & snowboard & avg &
    bear & camel & cows & dog & avg \\
    \midrule
    SPT$^+$       & 13.95 & 21.89 & 5.53 & 19.30 & 15.17 &
    \con & \con & \con & \con & \con \\
    NPR$^+$       & 21.03 & 23.95 & 11.96 & 29.54 & 21.62 &
    18.26 & 17.46 & 14.62 & 25.66 & 19.00 \\
    Pinocchio$^+$  & 16.91 & 18.63 & 7.64 & 18.38 & 15.39 &
    18.27 & 13.80 & 12.58 & 18.67 & 15.83 \\
    \midrule
    Ours          & \textbf{13.87} & \textbf{13.90} & \textbf{5.13} & \textbf{18.07} & \textbf{12.74} & 
    \textbf{13.31} & \textbf{12.28} & \textbf{6.43} & \textbf{17.94}  & \textbf{12.49} \\
    \bottomrule
  \end{tabular}
  }
  \vspace{-15pt}
\end{table*}

\paragraph{\textbf{Animal motion transfer comparisons}}
We further report animal motion transfer results in the DT4D-Quadrupeds and DT4D-Others columns of Tab.~\ref{tab:quantitative}. For reconstruct-and-retarget baselines, we use \cite{lyu2024animer} to reconstruct SMAL~\cite{zuffi20173dsmal} template pose sequences from the source videos.
We omit SPT$^+$ for animals because pretrained weights are not available for non-human shapes.
NPR$^+$ is omitted on DT4D-Others since it requires pretrained pose manifolds, which are not provided for non-quadruped animal categories.

For DT4D-Others, which includes categories beyond common parametric templates such as bird or whale, we compare our method against Transfer4D and Pinocchio$^+$.
For Pinocchio$^+$ on these categories, we provide a ground-truth source mesh sequence as input, bypassing the 2D video-to-3D reconstruction stage to avoid severe error propagation from an intermediate reconstruction pipeline.
As shown in Tab.~\ref{tab:quantitative}, MorphGS shows lower PMD and FVMD than both baselines, demonstrating effective motion transfer performance on non-standard categories that remain relatively underexplored in the motion retargeting literature.

\begin{figure*}[t]
    \centering
    \includegraphics[width=1.0\linewidth,trim={0 0cm 0cm 0},clip]{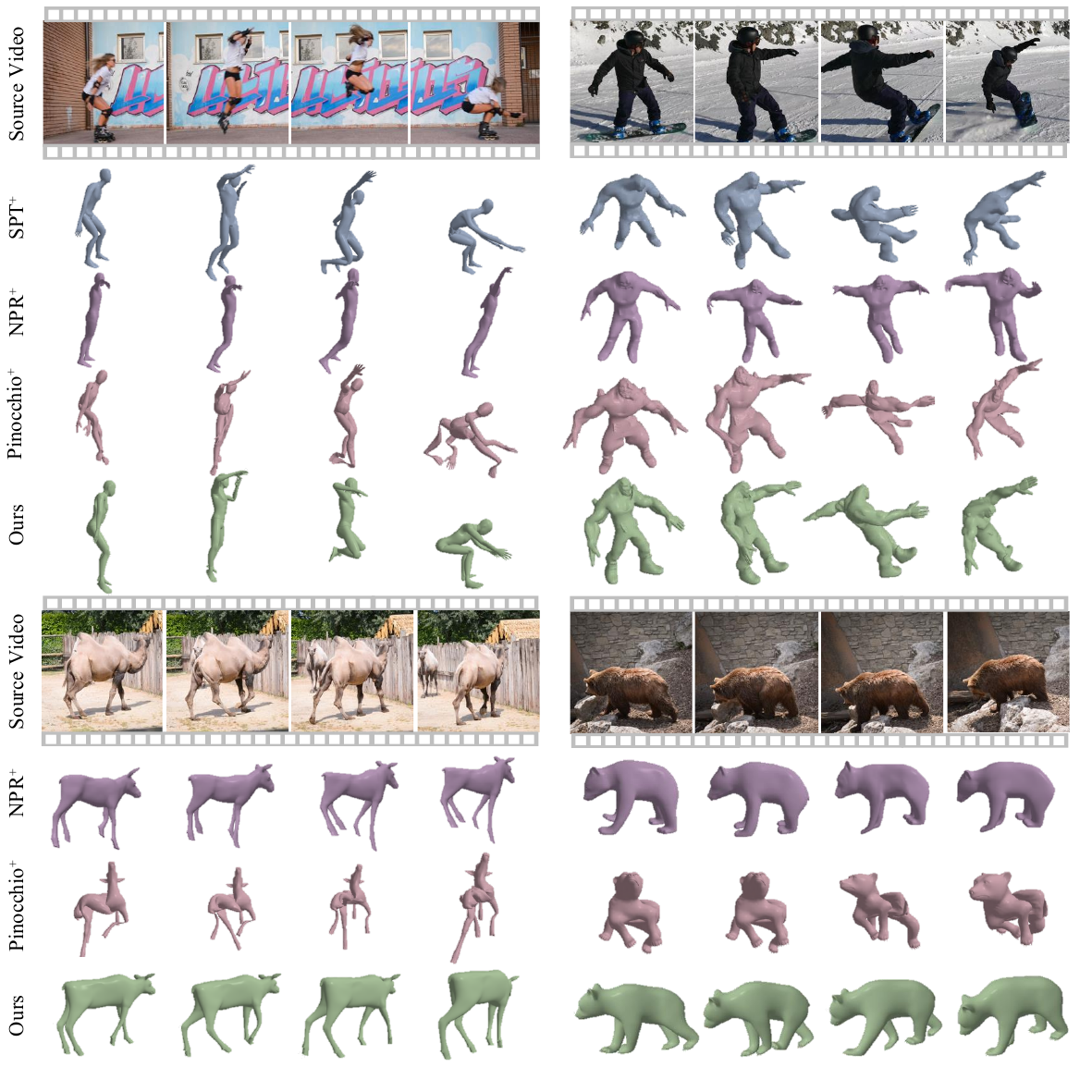}
    \vspace{-25pt}
    \caption{\footnotesize\textbf{Qualitative results on real-world videos.} We show motion transfer results on real-world videos.}
    \label{fig:qual_real}
    \vspace{-10pt}
\end{figure*}

\paragraph{\textbf{Real-world motion transfer comparisons}}
We evaluate on the real-world monocular videos curated from DAVIS~\cite{davis}, using them as motion sources.
Since ground-truth 3D mesh sequences are unavailable for such footage, we report FVMD~\cite{liu2024fr} as a perceptual measure of motion consistency rather than direct geometric error, and complement it with qualitative comparisons in Fig.~\ref{fig:qual_real} and the supplementary video.
We use target meshes from Mixamo~\cite{mixamo} for humanoids and DT4D~\cite{dt4d} for animals.
As summarized in Tab.~\ref{tab:quantitative-real}, MorphGS achieves a lower average FVMD than the compared baselines on both human and animal videos.
Qualitative results in Fig.~\ref{fig:qual_real} show that the posed targets track the motion in the input frames while maintaining the target's original structure.
These results indicate that our target-driven, morphology-adaptive optimization is effective on unconstrained monocular videos.

\begin{figure*}[t]
    \centering
    \includegraphics[width=\linewidth,trim=0 12.4cm 8.9cm 0,clip]{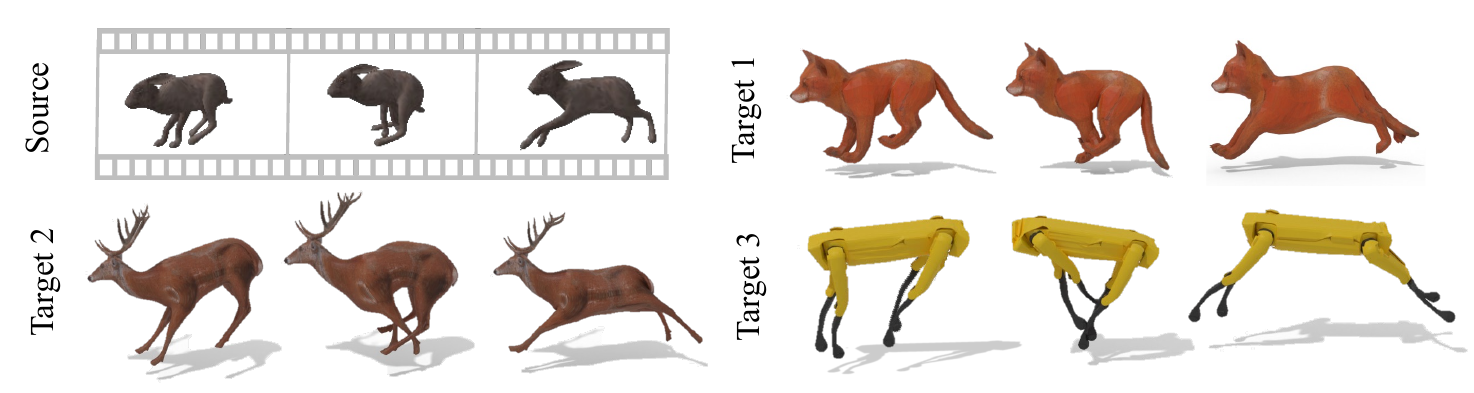}
    \vspace{-10pt}
    \caption{\textbf{Cross-category motion transfer.} A single rabbit source motion is retargeted onto structurally distinct targets. Each target keeps its own identity while following the source motion.}
    \label{fig:appendix_cross}
    \vspace{-5pt}
\end{figure*}

\paragraph{\textbf{Cross-category motion transfer}}
Figure~\ref{fig:appendix_cross} further demonstrates cross-category transfer, where a single rabbit running motion is retargeted to a fox, a deer, and a quadruped robot with substantially different body proportions and local geometries. 
This factorization of body proportions and local geometry enables MorphGS to reproduce the source running motion across targets while preserving target-specific morphology.

\subsection{Ablation Studies}
\label{sec:ablation}
In this section, we evaluate the key components of our framework by ablating the morphology parameterization and supervision signals.

\begin{figure}[t]
    \centering
    \begin{minipage}[t]{0.38\textwidth} 
        \vspace{0pt}
        \centering
        {\scriptsize
        \setlength{\tabcolsep}{3pt}
        \renewcommand{\arraystretch}{1.15}
        \captionof{table}{\textbf{Ablation study.} PMD (\( \times 10^3 \), ↓) and FVMD (\( \times 10^{-3} \), ↓) are averaged over 12 scenes from the Mixamo~\cite{mixamo} dataset.}
        \vspace{3pt}
        \label{tab:ablation}
        \begin{tabular*}{\linewidth}{@{\extracolsep{\fill}} l cc @{}}
            \toprule
            \textbf{Model} & \textbf{PMD} & \textbf{FVMD} \\
            \midrule
            \multicolumn{3}{@{}l}{\textit{Morphology parameterization}} \\
            \quad Na\"ive model  & 8.45 & 17.54 \\
            \quad Fixed morphology & 3.72 & 15.30 \\
            \quad + $\ell_b$ & 2.87 & 12.45 \\
            \quad + $\ell_b$, $s_\mathrm{global}$ & 1.99 & 9.55 \\
            \quad + $\ell_b$, $s_\mathrm{global}$, $\mathbf{o}_k$  & 1.91 & 8.82 \\
            \midrule
            \multicolumn{3}{@{}l}{\textit{Supervision losses}} \\
            \quad $\mathcal{L}_\mathrm{render}$ & 4.52 & 13.57 \\
            \quad + $\mathcal{L}_\mathrm{keyp}$ & 2.96 & 11.93 \\
            \quad + $\mathcal{L}_\mathrm{keyp}$, $\mathcal{L}_\mathrm{mv}$  & 1.91 & 8.82 \\
            \bottomrule
        \end{tabular*}}
    \end{minipage}
    \hfill 
    \begin{minipage}[t]{0.59\textwidth}
        \vspace{6pt}
        \centering
        \includegraphics[width=\linewidth, trim={0 5.5cm 14cm 0},clip]{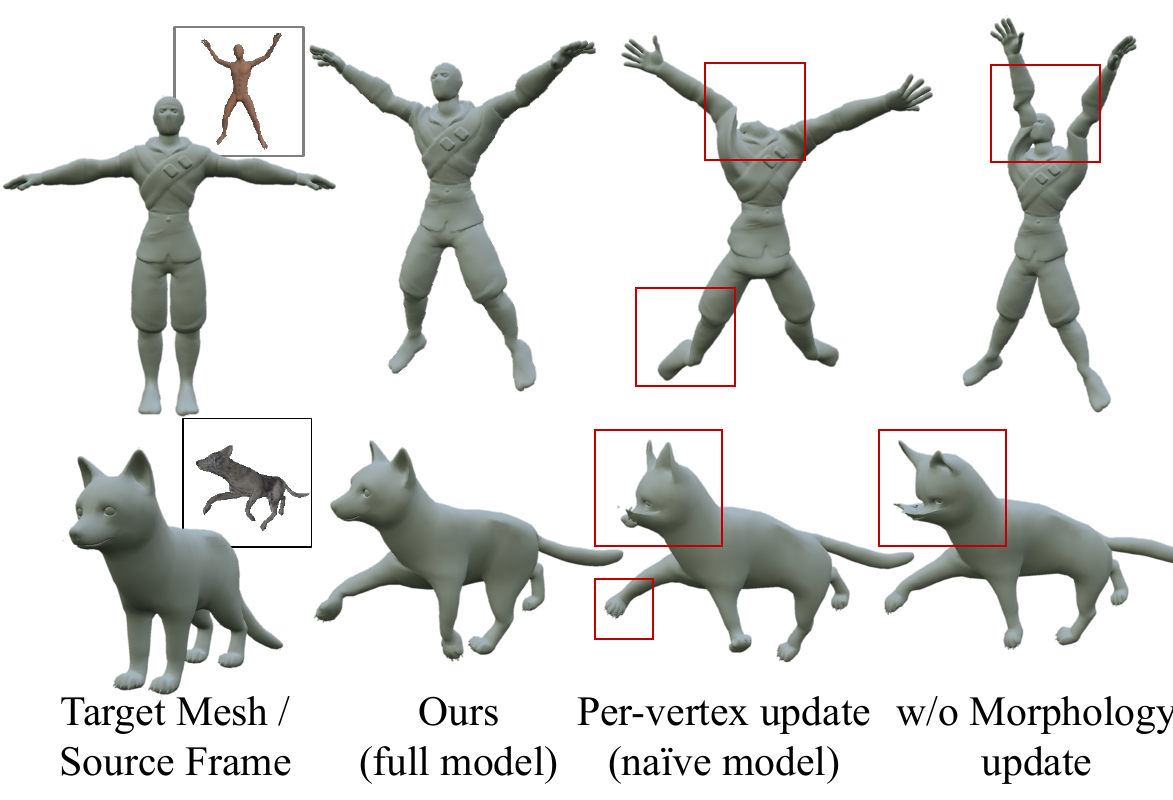}
        \captionof{figure}{\footnotesize\textbf{Effect of morphology parameterization.} Skeleton anchoring couples vertices to the rig, preventing structure-violating deformation.} 
        \label{fig:qual_ablation}
    \end{minipage}
\end{figure}

\paragraph{\textbf{Effects of morphology parameterization}} 
We first analyze the effect of our morphology parameterization by comparing it against a \emph{na\"ive model} that defines each canonical vertex  $\mathbf{v}_k\in\mathbb{R}^3$ as a free variable in the global coordinate, rather than as a local offset $\mathbf{o}_k$ relative to the skeleton-dependent anchor $\mathbf{p}_k$ (Eq.~\ref{eq:skeleton_anchor}). 
Without the anchor $\mathbf{p}_k$ coupling $\mathbf{v}_k$ to the bone lengths $\ell_b$, individual vertices can shift independently of the kinematic structure during optimization.
As shown in Tab.~\ref{tab:ablation} and Fig.~\ref{fig:qual_ablation}, this leads to significant performance degradation (PMD 8.45, FVMD 17.54).

We further ablate individual morphology parameters by progressively enabling them. Without morphology adaptation (fixed morphology), the optimizer compensates for the shape gap through incorrect joint rotations, degrading pose accuracy (PMD 3.72).
Enabling bone lengths $\ell_b$ and global scale $s_{\mathrm{global}}$ resolves skeletal proportion and scale mismatches, respectively, yielding substantial gains. Offsets $\mathbf{o}_k$ provide further refinement by capturing residual surface details.

\paragraph{\textbf{Effects of supervision losses}}
Rendering losses alone recover coarse motion but lack part-level precision, yielding a relatively high PMD of 4.52, since photometric and silhouette cues constrain the projected appearance and outline without resolving part-to-part correspondence. 
Adding dense 2D--3D keypoint supervision introduces explicit part-wise anchors, leading to a substantial reduction in PMD from 4.52 to 2.96. 
Synthesized-view guidance further improves robustness under self-occlusion by reducing monocular ambiguity, yielding the best performance with 1.91 PMD and 8.82 FVMD. 
Additional ablations on supervision losses are provided in the appendix.
\section{Conclusion and Future Work}
\label{sec:discussion}
In this study, we have presented MorphGS, a video-to-3D motion transfer framework. By formulating motion retargeting as an inverse-graphics optimization problem, our approach explicitly factorizes morphology and rig-driven pose parameters. This parameterization effectively addresses source-target shape misalignments, enabling motion transfer without relying on predefined parametric templates or intermediate 3D source reconstruction.

Although our morphology adaptation facilitates cross-identity motion transfer, the current framework assumes a compatible articulated topology between the source and target. This assumption inherently limits transfer across subjects with significantly different kinematic structures, such as differences in branch count or joint hierarchy. 
For future work, integrating physics-based motion priors could provide stronger constraints to resolve ambiguities under severe self-occlusions and reduce temporal artifacts. 
Moreover, explicitly modeling non-rigid soft-tissue deformations could further improve the expressiveness of the transferred motion. We hope our approach provides a useful step toward more flexible 3D content creation and cross-category motion understanding.

\section*{Acknowledgements}
Sung-Eui Yoon is the corresponding author. This work was supported by the National Research Foundation of Korea (NRF) grant funded by the Korea government (MSIT) (No. RS-2023-00208506), and in part by the Institute of Information \& communications Technology Planning \& Evaluation (IITP) grant funded by the Korea government (MSIT) (No.RS-2025-25443318, Physically-grounded Intelligence: A Dual Competency Approach to Embodied AGI through Constructing and Reasoning in the Real World).

%
%
\bibliographystyle{splncs04}
\bibliography{main}

\newpage
\appendix
\renewcommand{\thetable}{A\arabic{table}}
\setcounter{table}{0}

\renewcommand{\thefigure}{A\arabic{figure}}
\setcounter{figure}{0}

\setcounter{topnumber}{4}
\setcounter{totalnumber}{6}
\setcounter{dbltopnumber}{4}
\renewcommand{\topfraction}{0.95}
\renewcommand{\dbltopfraction}{0.95}
\renewcommand{\textfraction}{0.05}
\renewcommand{\floatpagefraction}{0.7}
\renewcommand{\dblfloatpagefraction}{0.7}

\title{MorphGS: Morphology-Adaptive Articulated 3D Motion Transfer from Videos}
\makeappendixtitle

In this appendix, we provide implementation details (Sec.~\ref{sec:appendix_impl}), extended ablation studies (Sec.~\ref{sec:appendix_ablation}), an analysis of morphology adaptation (Sec.~\ref{sec:appendix_morphology}), and an identifiability analysis of the proposed morphology parameterization (Sec.~\ref{sec:appendix_theory}). We further present additional real-world results that extend those in the main paper, along with full per-scene tables and 3D visualizations (Sec.~\ref{sec:appendix_perscene}).


\section{Implementation Details}
\label{sec:appendix_impl}

\subsection{Training objective and regularization}
Following Sec.~3.4 of the main paper, we optimize the following objective at each frame:
\begin{equation}
\mathcal{L}_t = \lambda_{\mathrm{render}}\bigl(\mathcal{L}_{\mathrm{render}} + \lambda_{\mathrm{mv}}\mathcal{L}_{\mathrm{mv}}\bigr) + \lambda_{\mathrm{keyp}}\mathcal{L}_{\mathrm{keyp}} + \mathcal{L}_{\mathrm{reg}},
\label{eq:training_objective}
\end{equation}
where \(
\mathcal{L}_{\mathrm{reg}}
=
\mathcal{L}_{\mathrm{smooth}}
+
\mathcal{L}_{\mathrm{chamf}}
+
\mathcal{L}_{\mathrm{trans}}
+
\mathcal{L}_{\mathrm{arap}}.
\) We define each term below.

\paragraph{\textbf{Temporal smoothness}}
To discourage abrupt changes over time, we further impose an $l1$ penalty on frame-to-frame differences in the joint motion parameters $\boldsymbol{\theta}_j^t$ and the root translation $\boldsymbol\delta_{\mathrm{root}}^{t}$:
\begin{equation}
\mathcal{L}_{\mathrm{smooth}}
= \lambda_{\mathrm{smooth}}
\biggl(
\sum_{j=1}^{J}\bigl\lVert \boldsymbol{\theta}_{j}^{t}-\boldsymbol{\theta}_{j}^{t-1}\bigr\rVert_{1}
+\bigl\lVert \boldsymbol\delta_{\mathrm{root}}^{t}-\boldsymbol\delta_{\mathrm{root}}^{t-1}\bigr\rVert_{1}
\biggr),
\end{equation}
where $\lambda_{\mathrm{smooth}}$ is the loss weight.

\paragraph{\textbf{2D Chamfer distance}}
Following Yao et al.~\cite{yao2025riggs}, we use a 2D Chamfer distance to align the deformed 3D skeleton with the source silhouette structure in the image plane. Specifically, let \(\mathcal{S}_t\) denote the 2D skeleton point set extracted from the source foreground mask \(\Omega_t\) by morphological thinning~\cite{zhang1984fast}. We sample 3D points \(c^t\) on the deformed skeleton and project them to the image plane with \(\pi_t\), yielding the projected point set \(\hat{\mathcal{S}_t}=\pi_t(c^t)\):
\begin{equation}
\mathcal{L}_{\mathrm{chamf}}
= \lambda_{\mathrm{chamf}}\,
\mathrm{CD}_{l_1}\bigl(\mathcal{S}_t, \hat{\mathcal{S}_t}\bigr),
\end{equation}
where \(\mathrm{CD}_{l_1}\) denotes the Chamfer distance~\cite{fan2017point} under the \(l_1\) norm, and $\lambda_{\mathrm{chamf}}$ is the corresponding loss weight.

\paragraph{\textbf{Transformation regularization}}
To discourage excessive joint rotation angles and root translations, we impose an \(\ell_1\) penalty on the per-frame root translation and joint rotation angles. Let \(\boldsymbol{\theta}_j^t = \alpha_j^t \hat{\mathbf{u}}_j^t\) denote the axis-angle pose parameter of joint \(j\) at frame \(t\), where \(\alpha_j^t\) is the rotation angle and \(\hat{\mathbf{u}}_j^t\) is the unit rotation axis.
\begin{equation}
\mathcal{L}_{\mathrm{trans}}
= \lambda_{\mathrm{trans}}
\frac{
\|\boldsymbol{\delta}_{\mathrm{root}}^{t}\|_{1}
+\sum_{j=1}^{J} |\alpha_{j}^{t}|
}{J}.
\end{equation}
Here, \(\lambda_{\mathrm{trans}}\) is the loss weight, \(\boldsymbol{\delta}_{\mathrm{root}}^{t}\) is the root translation at frame \(t\), and \(J\) is the number of joints.

\paragraph{\textbf{As-Rigid-As-Possible}}
We use an ARAP loss defined on canonical nearest-neighbor pairs. For each Gaussian \(k\), the neighborhood \(\mathcal{N}(k)\) is defined in the canonical pose using the canonical Gaussian centers \(\bar{\mathbf{v}}_k\). We then penalize deviations between the canonical distances \(\|\bar{\mathbf{v}}_k - \bar{\mathbf{v}}_{n}\|\) and the posed distances \(\|\mathbf{v}_k^t - \mathbf{v}_{n}^t\|\) for \(n\in\mathcal{N}(k)\), where \(\mathbf{v}_k^t\) denotes the center of Gaussian \(k\) at frame \(t\):

\begin{equation}
  \mathcal{L}_{\mathrm{arap}} = \lambda_{\mathrm{arap}}
  \frac{1}{K} \sum_{k=1}^{K} \frac{1}{|{\mathcal{N}(k)}|}
  \sum_{n \in \mathcal{N}(k)}
  \log\!\Big(1 + \big| \|\bar{\mathbf{v}}_k - \bar{\mathbf{v}}_{n}\|
  - \|\mathbf{v}_k^t - \mathbf{v}_{n}^t\| \big|\Big),
\end{equation}
We use the \(\log(1+\cdot)\) form to reduce the contribution of larger discrepancies.

\paragraph{\textbf{Loss weights}}
For Eq.~\ref{eq:training_objective}, we use \(\lambda_{\mathrm{render}}=1.0\), \(\lambda_{\mathrm{mv}}=0.25\), and \(\lambda_{\mathrm{keyp}}=0.006\).
For motion regularization, we use \(\lambda_{\mathrm{trans}}=0.005\), \(\lambda_{\mathrm{smooth}}=0.01\), \(\lambda_{\mathrm{chamf}}=0.001\), and \(\lambda_{\mathrm{arap}}=1.0\).


\subsection{Model and optimization details}
\paragraph{\textbf{Pose network}}  
The pose network is an MLP-based motion field following Uzolas et al.~\cite{uzolas2023template}. A scalar time input is encoded with a sinusoidal positional embedding and mapped to a 30D latent code by a small time encoder. This code is processed by an 8-layer MLP with 256 hidden units and a skip connection. Two linear heads predict the global 3D translation and per-joint axis-angle rotations. These are converted to joint \(SE(3)\) transforms via Rodrigues' formula and skeletal kinematics, and applied using linear blend skinning.

\paragraph{\textbf{Optimization details}}
We use learning rates of 0.005, 0.01, and 0.001 for the motion network, global scale, and bone lengths, respectively. 
Bone lengths are enabled after the initial global-alignment stage, starting at iteration 500, and optimized with a learning rate of 0.001. 
A step decay of 0.1 is applied every 1K iterations thereafter. 
For Gaussian attributes including offsets, we use the default learning-rate settings of 3D Gaussian Splatting~\cite{kerbl20233d}.

\subsection{Camera and synthesized-view setup}
\paragraph{\textbf{Camera setup}}
In all experiments, we render the posed target from a fixed reference camera chosen from the predefined canonical views of the target mesh, with intrinsics kept fixed throughout optimization. 
Instead, frame-wise source–target alignment is modeled by a per-frame root transform in \(SE(3)\). The same setup is used for both synthetic and real videos regardlessly.
For synthesized-view guidance, auxiliary views are rendered by rotating the fixed reference camera around the object center by the prescribed azimuth offset~\cite{yao2025sv4d}, while keeping the intrinsics unchanged.
\section{Extended Ablation Studies}
\label{sec:appendix_ablation}

We extend the ablation studies in Sec.~4.4 of the main paper
with additional analysis on design choices. All experiments are
conducted on the Mixamo dataset unless otherwise noted.

\begin{table}[t]
  \centering
  \begin{minipage}[t]{0.32\linewidth}
    \captionsetup{type=table,justification=justified,singlelinecheck=false}
    \caption{\footnotesize\textbf{Effect of semantic feature extractors.} PMD ($\times 10^{3}$, $\downarrow$) and FVMD ($\times 10^{-3}$, $\downarrow$) for different frozen feature extractors show that Geo-Aware yields the best overall results.}
    \vspace{-1pt}
    \label{tab:appendix_ablate_tlfr}
    {\fontsize{8pt}{9.5pt}\selectfont
    \setlength{\tabcolsep}{2pt}
    \begin{tabularx}{\linewidth}{@{\extracolsep{\fill}} X c c @{}}
      \toprule
      Method                   & PMD & FVMD \\
      \midrule
      DINOv2~\cite{oquab2023dinov2} & 2.90 & 9.45 \\
      DINOv3~\cite{simeoni2025dinov3}  & 2.19 & 9.42 \\
      Geo-Aware~\cite{zhang2024telling}        & \textbf{1.91} & \textbf{8.82} \\
      \bottomrule
    \end{tabularx}
    }
  \end{minipage}
  \hfill
  \begin{minipage}[t]{0.30\textwidth}
    \captionsetup{type=table,justification=justified,singlelinecheck=false}
    \caption{\footnotesize \textbf{Effect of correspondence density.} PMD ($\times 10^{3}$, $\downarrow$) and FVMD ($\times 10^{-3}$, $\downarrow$) remain stable as the retained ratio of sampled correspondences decreases. }
    \label{tab:appendix_ablate_keyp}
    \parbox[t][2.6cm][t]{\linewidth}
    {\fontsize{8pt}{9.5pt}\selectfont
    \setlength{\tabcolsep}{2pt}
    \begin{tabularx}{\linewidth}{@{\extracolsep{\fill}} X c c @{}}
      \toprule
      Ratio (\%) & PMD & FVMD \\
      \midrule
      100   & \textbf{1.91} & \textbf{8.82} \\
      50  & 1.94 & 9.05 \\
      30  & 1.95 & 9.14 \\
      10  & 1.96 & 9.36 \\
      \bottomrule
    \end{tabularx}
    }
  \end{minipage}
  \hfill
    \begin{minipage}[t]{0.34\linewidth}
        \captionof{table}{{\footnotesize\textbf{Analysis on synthesized-view supervision.} PMD ($\times 10^{3}$, $\downarrow$) and FVMD ($\times 10^{-3}$, $\downarrow$) improve consistently as the number of synthesized views increases.}}
        \label{tab:appendix_ablate_mv}
        \parbox[t][2.6cm][t]{\linewidth}
        {\fontsize{8pt}{9.5pt}\selectfont
        \setlength{\tabcolsep}{3pt} 
        \begin{tabularx}{\linewidth}{X c c}
        \toprule
        \#Synth. views & PMD & FVMD \\
        \midrule
          0 (w/o $\mathcal{L}_{mv}$)  & 2.96 & 11.93 \\
          1 & 2.27 & 11.21 \\
          2 & 2.18 & 9.41 \\
          4 & 1.91 & 8.82 \\
          8 & \textbf{1.85} & \textbf{7.97} \\
          \bottomrule
        \end{tabularx}
        }
    \end{minipage}
    \vspace{-10pt}
\end{table}

\paragraph{\textbf{Geometry-aware semantic features}}
Distinguishing semantically similar but geometrically different parts, such as left and right limbs, is important for accurate correspondence matching. We therefore use a pretrained geometry-aware semantic feature extractor (Geo-Aware)~\cite{zhang2024telling}. 
We compare Geo-Aware, DINOv2~\cite{oquab2023dinov2}, and DINOv3~\cite{simeoni2025dinov3} while keeping the rest of the optimization pipeline fixed. 
All feature extractors are used frozen.
Tab.~\ref{tab:appendix_ablate_tlfr} shows that Geo-Aware achieves the lowest PMD and FVMD among the compared variants, indicating that features trained to resolve left-right ambiguity are better suited to our motion learning setting.

\paragraph{\textbf{Keypoint density}}
We further vary the density of keypoint supervision by randomly retaining a fraction of the 1K sampled correspondences per frame. 
As shown in Tab.~\ref{tab:appendix_ablate_keyp}, reducing the retained ratio from 100\% to 10\% causes only a modest degradation, with PMD increasing from 1.91 to 1.96 (+2.6\%) and FVMD from 8.82 to 9.36 (+6.1\%). 
This indicates that the sampled correspondences mainly constrain coarse part-level alignment rather than fine-grained geometry. Once this alignment is provided, further increasing the correspondence density has a limited effect on the final result.

\paragraph{\textbf{Number of synthesized views}}
Tab.~\ref{tab:appendix_ablate_mv} reports the effect of varying the number of synthesized views $N$ used in $\mathcal{L}_{\mathrm{mv}}$ (Eq.~13). 
PMD and FVMD decrease monotonically as $N$ increases from 0 to 8, confirming that additional viewpoints provide complementary supervision, with diminishing returns beyond $N=4$. 
We use $N=4$ as the default, as increasing to $N=8$ yields only a modest 
further improvement (PMD: 1.91 $\rightarrow$ 1.85, FVMD: 8.82 $\rightarrow$ 7.97) 
while increasing the cost of novel-view synthesis.
Notably, even without multi-view supervision ($N=0$), MorphGS achieves PMD of 2.96 and FVMD of 11.93, already matching SPT\textsuperscript{+} in PMD and outperforming it in FVMD (12.56), indicating that the rendering and keypoint losses alone provide a competitive baseline.
We further extend the monocular ($N=0$) versus multi-view ($N=4$) comparison to every evaluated synthetic dataset split in Tab.~\ref{tab:appendix_monocular}, where MorphGS remains competitive with or surpasses the best baseline.

\begin{table}[t]
    \centering
    \captionsetup{font=footnotesize}
    \caption{\textbf{Extended monocular ablation analysis.} Best baseline indicates SPT$^+$ for Mixamo, NPR$^+$ for DT4D-Quadrupeds, and Transfer4D for DT4D-Others.}
    \label{tab:appendix_monocular}
    \fontsize{8pt}{9.5pt}\selectfont
    \setlength{\tabcolsep}{0pt}
    \newcolumntype{N}{>{\centering\arraybackslash}p{0.11\linewidth}}
    \begin{tabular*}{\linewidth}{@{\extracolsep{\fill}}l *{6}{N}}
        \toprule
        & \multicolumn{2}{c}{Mixamo} & \multicolumn{2}{c}{DT4D-Quadrupeds} & \multicolumn{2}{c}{DT4D-Others} \\
        \cmidrule(lr){2-3} \cmidrule(lr){4-5} \cmidrule(lr){6-7}
        \textbf{Method} & PMD & FVMD & PMD & FVMD & PMD & FVMD \\ \midrule
        Best baseline* & 2.96 & 12.56 & 3.28 & 16.32 & 7.16 & 17.52 \\
        \midrule
        \textbf{Ours w/o $\mathcal{L}_{mv}$} & 2.96 & 11.93 & 2.29 & 11.75 & 3.16 & 13.15 \\
        \textbf{Ours} & \textbf{1.91} & \textbf{8.82} & \textbf{1.33} & \textbf{8.81} & \textbf{1.95} & \textbf{12.83} \\
        \bottomrule
    \end{tabular*}
\end{table}

\begin{figure*}[t]
    \centering
    \begin{minipage}[t]{0.53\linewidth}
    \captionsetup{type=table,justification=justified,singlelinecheck=false}
        \caption{\footnotesize\textbf{Effect of target rig initialization.} Results on DT4D-Quadrupeds with different auto-rigging methods used to initialize the target rig. PMD ($\times 10^{3}$, $\downarrow$) and FVMD ($\times 10^{-3}$, $\downarrow$) vary with the choice of rig initializer.}
        \label{tab:appendix_ablate_rigging}
        {\fontsize{8pt}{9.5pt}\selectfont
        \setlength{\tabcolsep}{2pt}
        \renewcommand{\arraystretch}{1.05}
        \begin{tabularx}{\linewidth}{@{\extracolsep{\fill}} X *{2}{c} @{}}
          \toprule
          {Method} & {PMD} & {FVMD}  \\
          \midrule
          MorphGS + Pinocchio~\cite{baran2007automatic}   & 1.49 & 9.79  \\
          MorphGS + RigNet~\cite{xu2020rignet}            & 1.33 & 8.81  \\
          MorphGS + MagicArticulate~\cite{song2025magicarticulate}   & 1.21 & 8.94 \\
          MorphGS + Puppeteer~\cite{song2025puppeteer} & \textbf{1.14} & \textbf{8.62}  \\
          \bottomrule
        \end{tabularx}
        }
  \end{minipage}
    \hfill
    \begin{minipage}[t]{0.45\linewidth}
        \captionof{table}{{\footnotesize\textbf{Effect of pose parameterization.} Comparison between direct per-frame pose optimization and a time-conditioned pose network. The pose network yields lower PMD ($\times 10^{3}$), FVMD ($\times 10^{-3}$), and Jitter (km/s$^{3}$) than per-frame pose optimization.}}
        \label{tab:appendix_ablate_pose}
        {\fontsize{8pt}{9.5pt}\selectfont
        \setlength{\tabcolsep}{3pt} 
        \begin{tabularx}{\linewidth}{X c c c}
        \toprule
        Method & PMD & FVMD & Jitter \\
        \midrule
          Per-frame Pose  & 2.60 & 12.98 & 0.15 \\
          Pose Network & \textbf{1.91} & \textbf{8.82} & \textbf{0.09} \\
          \bottomrule
        \end{tabularx}
        }
    \end{minipage}
\end{figure*}

\paragraph{\textbf{Auto-rigging methods}}
We ablate the target rig initialization used in MorphGS by replacing 
the off-the-shelf auto-rigging tool~\cite{baran2007automatic,xu2020rignet,song2025magicarticulate,song2025puppeteer}
while keeping the rest of the optimization unchanged.
We report this comparison on DT4D-Quadrupeds~\cite{dt4d}, focusing on the setting where target rig initialization is required and all compared rigging methods are applicable.
As shown in Tab.~\ref{tab:appendix_ablate_rigging}, MorphGS is applicable with multiple auto-rigging methods, while stronger rig initialization leads to better PMD and FVMD. 

\paragraph{\textbf{Pose network vs.\ per-frame pose optimization}}
Tab.~\ref{tab:appendix_ablate_pose} compares direct per-frame pose 
optimization against our time-conditioned pose network (Sec.~3.3).
We additionally report \emph{Jitter}~\cite{PIPCVPR2022}, the mean 
jerk of all body joints in global space (km/s$^3$), 
which reflects the temporal smoothness of the recovered motion.
The pose network outperforms per-frame optimization across all three 
metrics (PMD: 1.91 vs.\ 2.60, FVMD: 8.82 vs.\ 12.98, Jitter: 0.09 
vs.\ 0.15). The gain is consistent with the MLP's implicit temporal 
continuity, which encourages motion consistency across adjacent frames 
and suppresses high-frequency artifacts, as reflected in the lower 
Jitter score.
\section{Analysis of Morphology Adaptation}
\label{sec:appendix_morphology}

\begin{table*}[t]
    \centering
    \caption{\footnotesize\textbf{Quantitative evaluation of adapted skeletons on two character pairs.} MPJPE is measured in the rest pose. PCK@0.1 and PCK@0.2 use thresholds of 10\% and 20\% of the characteristic length ($L_{\mathrm{char}}=\sqrt[3]{V_{\mathrm{GT}}}$), respectively.}
    \label{tab:appendix_skeletal}
    \vspace{4pt}
    {\fontsize{8pt}{9.5pt}\selectfont
    \renewcommand{\arraystretch}{1.05}
    \begin{tabular*}{\linewidth}{@{\extracolsep{\fill}}l *{6}{c}@{}}
    \toprule
    Method 
    & \multicolumn{3}{c}{Mannequin $\rightarrow$ Ortiz} 
    & \multicolumn{3}{c}{Mannequin $\rightarrow$ Pumpkinhulk} \\
    \cmidrule(lr){2-4} \cmidrule(lr){5-7}
    & MPJPE $\downarrow$ & PCK@0.1 $\uparrow$ & PCK@0.2 $\uparrow$
    & MPJPE $\downarrow$ & PCK@0.1 $\uparrow$ & PCK@0.2 $\uparrow$ \\
    \midrule
    Pinocchio~\cite{baran2007automatic} & 0.1013 & 0.6316 & 0.8947 & 0.1280 & 0.3684 & 0.8947 \\
    Ours      & \textbf{0.0736} & \textbf{0.7368} & \textbf{0.9868} & \textbf{0.0790} & \textbf{0.6316} & \textbf{0.9605} \\
    \bottomrule
    \end{tabular*}
    }
\end{table*}

\begin{figure}[t]
    \centering
    \includegraphics[width=1.0\linewidth,trim=0 11cm 0cm 0,clip]{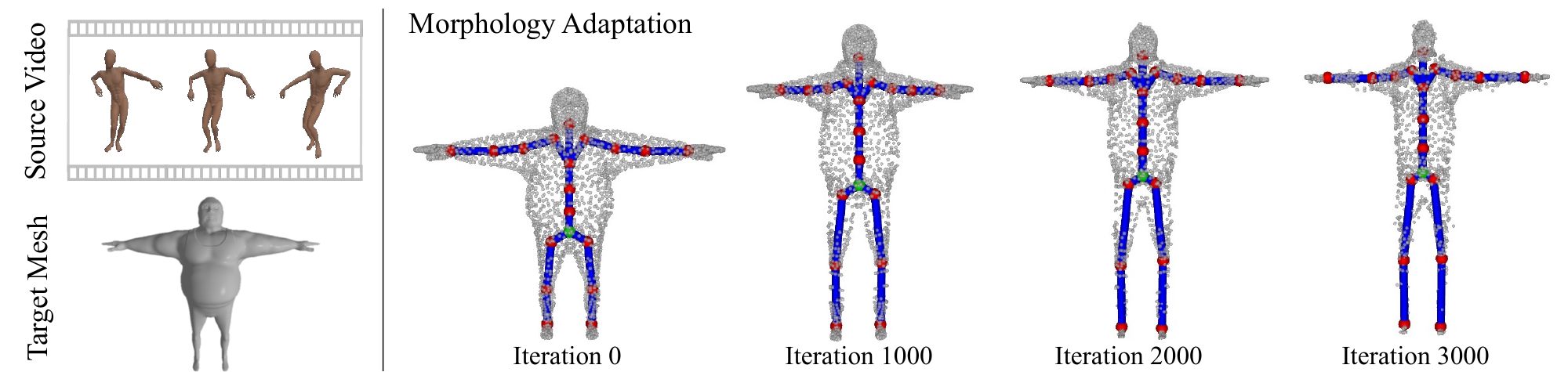}
    \caption{\textbf{Visualization of morphology adaptation.} During optimization, the global scale, bone lengths, and offset vectors are updated from the target initialization to better match the source subject.}
    \label{fig:appendix_morph_vis}
\end{figure}

\begin{figure}[t]
    \vspace{-7pt}
    \centering
    \includegraphics[width=0.8\linewidth,trim=0 2.9cm 3.2cm 0,clip]{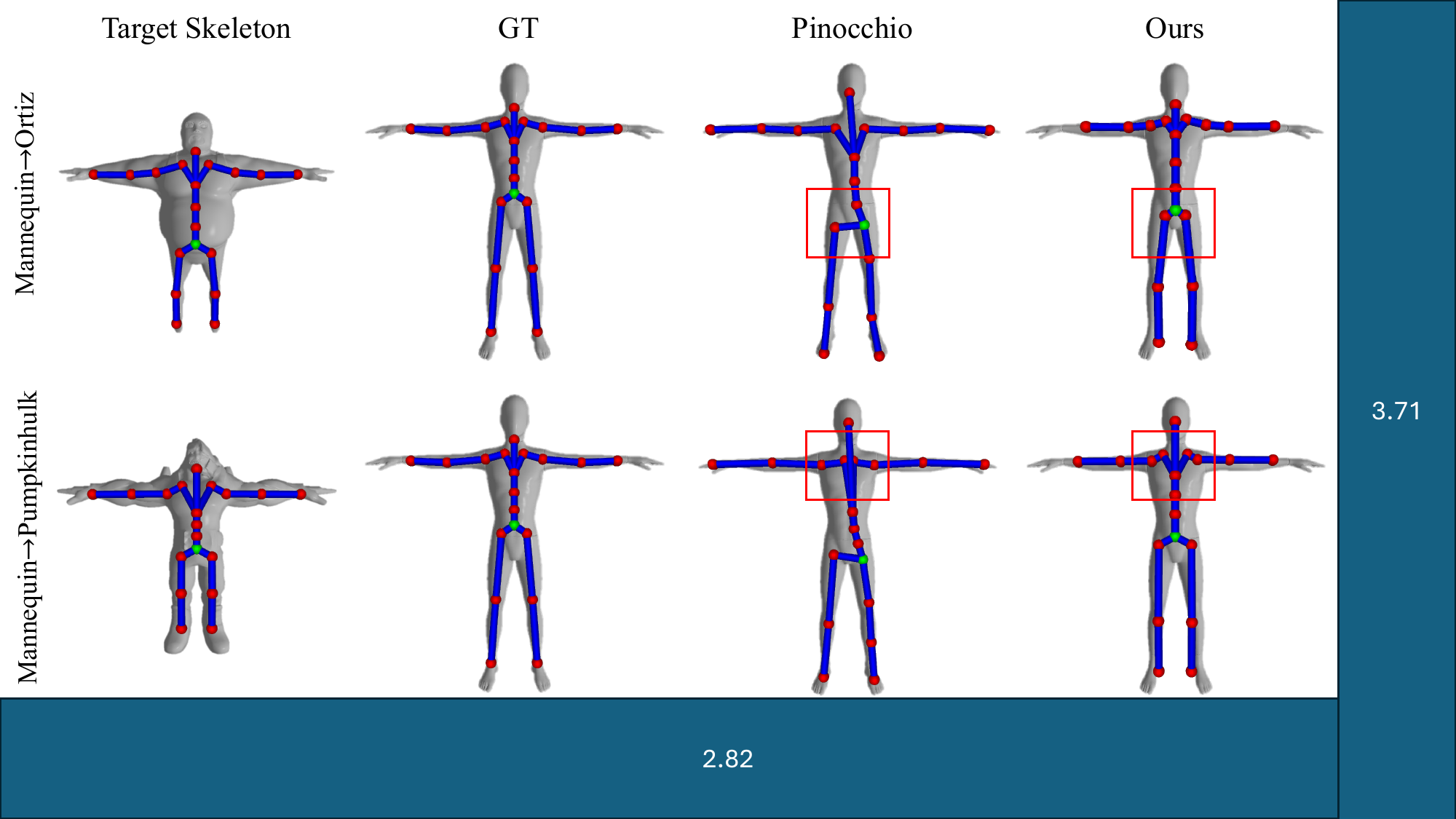}
    \caption{\textbf{Qualitative comparison of adapted skeletons in the rest pose.} Joint positions and bones are shown as colored spheres overlaid on the target mesh for two character pairs: Mannequin \(\rightarrow\) Ortiz and Mannequin \(\rightarrow\) Pumpkinhulk. Red boxes highlight the region where our adapted skeleton is more closely aligned with the ground-truth structure.}
    \label{fig:appendix_skeleton_qual}
\end{figure}

We analyze whether the optimized morphology parameters yield geometrically meaningful skeletal adaptation. Fig.~\ref{fig:appendix_morph_vis} shows morphology adaptation during optimization.
The global scale, bone lengths, and offset vectors are updated from the target initialization.

For evaluation, we measure how closely the adapted target skeleton matches the ground-truth skeleton in the rest pose. 
We compare against Pinocchio~\cite{baran2007automatic}, a geometry-driven auto-rigging baseline that fits the target skeleton directly to the source mesh. In contrast, our method adapts the target skeleton solely from the source video, without access to the source 3D geometry.
We report normalized Mean Per-Joint Position Error (MPJPE, $\downarrow$) and Percentage of Correct Keypoints (PCK, $\uparrow$)~\cite{rhodin2018learning} between the adapted skeleton and the ground-truth skeleton. PCK@0.1 and PCK@0.2 use thresholds of 10\% and 20\% of the characteristic length \(L_{\mathrm{char}}=\sqrt[3]{V_{\mathrm{GT}}}\), where \(V_{\mathrm{GT}}\) denotes the volume of the ground-truth 3D character mesh. For each character pair, the results are averaged over five motion sequences.


As shown in Tab.~\ref{tab:appendix_skeletal} and Fig.~\ref{fig:appendix_skeleton_qual}, our method yields lower MPJPE and higher PCK than Pinocchio on both character pairs. This indicates that the adapted skeleton is more closely aligned with the ground-truth skeleton, while not requiring the source mesh.

\section{Theoretical Analysis of Morphology Parameterization}
\label{sec:appendix_theory}
In this section, we provide an additional theoretical motivation for our
morphology parameterization and the staged optimization strategy in Sec.~4.1.
By coupling surface geometry to the kinematic chain through a small set of
structured parameters, the monocular optimization problem becomes
well-constrained: during the early optimization stage where offsets
$\{\mathbf{o}_k\}$ remain fixed (optimization stage 1), the skeletal
proportions are identifiable up to a global scale under assumptions on the
observed motion. We adopt the notation from the main paper (Eq. 3--6).

\subsection{Setup}
\label{sec:theory:setup}
Consider a monocular video $\{I_t\}_{t=1}^{T}$ of an articulated
object observed under weak-perspective projection.  The object consists of
$B$ rigid bones forming a kinematic tree $\mathcal{T}$.  We assume:
\begin{itemize}
  \item[$(\mathcal{A}_1)$]  \textbf{Piecewise rigidity.}
    Each bone is a rigid body, and its attached surface vertices maintain
    fixed local-frame positions across all frames. 
    This serves as an idealized approximation of the rig-coupled surface
    model used in the main method.
  \item[$(\mathcal{A}_2)$]  \textbf{Non-degenerate motion.}
    Each bone undergoes sufficiently rich motion across the sequence,
    including rotations about at least two linearly independent axes,
    and adjacent bones exhibit non-degenerate relative motion. The
    motion is temporally continuous, so depth ordering does not change
    discontinuously across consecutive frames.
  \item[$(\mathcal{A}_3)$] \textbf{Weak-perspective projection with fixed camera.}
    The camera coordinate frame is fixed. The 2D projection of
    $\mathbf{x}=[x,y,z]^\top$ at frame $t$ is
    $\boldsymbol{\pi}_t(\mathbf{x})=s_f^t[x,y]^\top + \mathbf{t}^t$,
    where $s_f^t > 0$ is a per-frame scale factor and $\mathbf{t}^t \in \mathbb{R}^2$ is a 2D translation.
\end{itemize}
With the local offsets $\{\mathbf{o}_k\}$ fixed from the target rig,
the canonical vertex position is derived from morphology parameters:
\[
\bar{\mathbf{v}}_k
= s_{\mathrm{global}}
\bigl(\mathbf{p}_k(\{\ell_b\}) + \mathbf{o}_k\bigr),
\]
as defined in Eq.~5--6 of the main paper. The unknowns thus reduce to
the global scale $s_{\mathrm{global}}$, bone lengths $\{\ell_b\}$, and
per-frame joint rotations $\{\boldsymbol{\theta}_j^t\}$, which in turn
determine the per-bone orientations $\{\mathbf{R}_b^t\}$ through
forward kinematics (Eq.~7--8).

\subsection{Identifiability Statement}

Our optimization begins from the target character's bone proportions
and adjusts them to best explain the source video observations.
Under the idealized assumptions above, the structured morphology
parameterization substantially reduces ambiguity compared with
unconstrained vertex-wise deformation. The following proposition states
that, in this early stage with fixed offsets, the skeletal morphology
is generically identifiable up to a single global scale.

\begin{proposition}[Identifiability of Skeletal Morphology]
\label{prop:ident}
Under $\mathcal{A}_1$--$\mathcal{A}_3$, suppose that each bone has at least four
non-coplanar associated surface points, and that these points are
observed across sufficiently many frames with non-degenerate motion.
Then the morphology parameters
$(s_{\mathrm{global}},\{\ell_b\})$ and the per-frame bone orientations
$\{\mathbf{R}_b^t\}$ are determined by the 2D observations
up to a single global scale factor. In particular, all bone-length
ratios $\ell_b/\ell_{b'}$ are uniquely determined.
\end{proposition}

\subsection{Proof Sketch of Proposition~\ref{prop:ident}}

We now provide a proof sketch of Proposition~\ref{prop:ident},
explaining how the structured morphology parameterization reduces the
ambiguity of the monocular reconstruction problem.

\paragraph{\textbf{Step 1: Per-bone structure recovery up to a similarity ambiguity}}
Under $\mathcal{A}_1$, each bone $b$ is associated with a set of surface
points whose relative configuration is fixed over time. For such points,
the weak-perspective observation at frame $t$ can be written as
\begin{equation}
\mathbf{u}_k^t
=
s_f^t\,\mathbf{R}_{b,2\times 3}^t \mathbf{v}_k + \mathbf{t}_b^t,
\end{equation}
where $s_f^t > 0$ is the per-frame projection scale shared across all bones,
$\mathbf{R}_{b,2\times 3}^t$ denotes the first two rows of the
bone pose in the camera frame, and $\mathbf{t}_b^t \in \mathbb{R}^2$
collects the projected translation term. Subtracting the per-frame
centroid removes translation, yielding:
\begin{equation}
\tilde{\mathbf{u}}_k^t
=
s_f^t\,\mathbf{R}_{b,2\times 3}^t \tilde{\mathbf{v}}_k .
\end{equation}
Let
\begin{equation}
\mathbf{S}_b =
\begin{bmatrix}
\tilde{\mathbf{v}}_1 & \cdots & \tilde{\mathbf{v}}_{P_b}
\end{bmatrix}
\in \mathbb{R}^{3\times P_b}
\end{equation}
collect the centered 3D point coordinates of bone $b$, and let:
\begin{equation}
\mathbf{M}_b =
\begin{bmatrix}
s_f^1\,\mathbf{R}_{b,2\times3}^1 \\
s_f^2\,\mathbf{R}_{b,2\times3}^2 \\
\vdots \\
s_f^T\,\mathbf{R}_{b,2\times3}^T
\end{bmatrix}
\in \mathbb{R}^{2T\times3}
\end{equation}
stack the scaled projected rotations across frames. Then stacking the centered
2D observations over all frames gives the measurement matrix:
\begin{equation}
\tilde{\mathbf{W}}_b = \mathbf{M}_b \mathbf{S}_b,
\end{equation}
where $\tilde{\mathbf{W}}_b \in \mathbb{R}^{2T\times P_b}$.
By construction, $\mathrm{rank}(\tilde{\mathbf{W}}_b)
\le 3$. Under the stated assumptions, this rank is generically equal to
$3$. This is because the associated surface points span a three-dimensional local
configuration, and the motion is sufficiently rich across frames
according to $\mathcal{A}_2$.
It then follows from standard scaled-orthographic rigid structure-from-motion
factorization arguments~\cite{tomasi1992shape,poelman1997paraperspective} that
$\tilde{\mathbf{W}}_b$ determines the 3D point configuration of bone
$b$ and its per-frame pose sequence up to a bone-specific similarity
ambiguity. Concretely, for each bone one may recover a shape basis and
motion matrices that are equivalent to the true ones up to an invertible
$3\times 3$ transform. After enforcing the metric
constraints (i.e., each row pair of $\mathbf{M}_b$ consists of two orthogonal vectors of equal norm $s_f^t$), the remaining ambiguity reduces to a similarity
transform consisting of a scale, a 3D rigid change of coordinates, and
possibly a reflection. Temporal continuity helps exclude inconsistent
frame-to-frame sign flips, but a per-bone gauge ambiguity still remains
at this stage. Therefore, Step~1 recovers each bone only up to an
independent similarity class, not yet in a common global frame.

\paragraph{\textbf{Step 2: Scale and orientation unification via joint constraints}}
Adjacent bones are not independent because neighboring bones share joints. If bones $b$ and $b'$ share a joint $j$, then their
reconstructed motions must be consistent with the same joint trajectory
across frames.
These shared-joint constraints couple the per-bone similarity
ambiguities left unresolved in Step~1. Under non-degenerate relative
motion~($\mathcal{A}_2$), they fix both the relative scale and relative
orientation between adjacent bones, since an incorrect relative scale or
an inconsistent reflection would cause the joint trajectories inferred
from the two bones to disagree over time.
Applying this argument recursively along the kinematic tree unifies all
per-bone scales into a single global scale~$\alpha$ and all per-bone
reflections into at most a single global reflection. The per-frame
projection scale $s_f^t$ is shared across all bones and cancels in
these inter-bone constraints. Finally, temporal
continuity~($\mathcal{A}_2$) resolves the remaining global reflection,
as only one of the two choices produces a depth trajectory that varies
continuously over time.

\paragraph{\textbf{Step 3: Bone length recovery}}
Once the joint trajectories are expressed in a common 3D frame up to the
single global scale $\alpha$, the length of bone $b$ is given by the
distance between its parent and child joints. Let $\mathbf{j}^t(j)$
denote the recovered 3D position of joint $j$ at frame $t$. Then
\begin{equation}
\hat{\ell}_b
=
\left\|
\mathbf{j}^t(j_{\mathrm{child}})
-
\mathbf{j}^t(j_{\mathrm{parent}})
\right\|.
\end{equation}
Because this parent--child distance is invariant for a rigid bone,
$\hat{\ell}_b$ is independent of $t$ and satisfies
\begin{equation}
\hat{\ell}_b = \alpha\,\ell_b .
\end{equation}
Therefore, all bone-length ratios $\ell_b/\ell_{b'}$ are determined,
while the absolute lengths remain ambiguous only up to the single global
scale absorbed into $s_{\mathrm{global}}$.


\paragraph{\textbf{Step 4: Bone orientation and twist disambiguation}}
The preceding steps determine the joint trajectories and bone lengths in
a common 3D frame up to the single global scale. This is sufficient to
fix the bone axis of each segment, but not, in general, the rotation
about that axis. In other words, recovering orientation from joint
positions alone leaves a residual twist ambiguity.
This remaining ambiguity is resolved by the surface observations. Each
bone is associated with surface points whose local offsets remain fixed,
and these offsets generically include components perpendicular to the
bone axis. A twist rotation about the bone axis therefore moves such
off-axis points, which in turn changes their projected 2D locations.
Consequently, different twist angles produce different image
observations.
Therefore, under generic non-axisymmetric local surface geometry, the
surface observations determine the residual twist angle of each bone.
Together with Steps~1--3, this yields the per-frame bone orientations
$\{\mathbf{R}_b^t\}$ uniquely, while the morphology parameters
$(s_{\mathrm{global}}, \{\ell_b\})$ are determined up to the single
global scale already accounted for above. \qed

\subsection{Practical Implications}
\label{sec:theory:practical}


\paragraph{\textbf{Effect of structured parameterization.}}
In a na\"ive formulation, each vertex $\bar{\mathbf{v}}_k \in \mathbb{R}^3$ is optimized independently, 
allowing individual vertices to drift freely regardless of the underlying skeletal structure.
Our parameterization instead decomposes each vertex into a skeleton-dependent anchor $\mathbf{p}_k(\{\ell_b\})$ 
and a local offset $\mathbf{o}_k$ (Eq.~5--6), 
coupling vertex positions to the kinematic chain through bone lengths and skinning weights.
As a result, adjusting a single bone length coherently displaces all vertices attached to the affected limb, 
enforcing that shape changes remain structurally consistent across the body.
This kinematic coupling is what enables the identifiability result in Proposition~1: 
in the early stage where offsets are fixed, 
the skeletal proportions are determined up to a single global scale 
from monocular observations alone.

\paragraph{\textbf{Fixing offsets in the early stage.}}
Although our parameterization couples vertex positions to the kinematic chain, 
a potential ambiguity remains when offsets $\{\mathbf{o}_k\}$ and bone lengths are optimized simultaneously: 
a change $\ell_b \rightarrow \ell_b + \delta\ell_b$ shifts the anchor $\mathbf{p}_k$ by $\Delta\mathbf{p}_k$, 
which can be compensated by $\mathbf{o}_k \rightarrow \mathbf{o}_k - \Delta\mathbf{p}_k$, 
leaving the canonical vertex position $\mathbf{v}_k = \mathbf{p}_k + \mathbf{o}_k$ largely unchanged.
Fixing the offsets during the early optimization stage, 
while estimating only the skeletal parameters, 
removes this potential ambiguity 
and allows bone lengths to be determined primarily from the observed motion.
The offsets are released in the later stage 
once the skeleton is sufficiently constrained, 
to capture residual shape details beyond what the skeletal proportions alone can express.

\paragraph{\textbf{Role of offsets under morphology mismatch.}}
The identifiability analysis above relies on idealized assumptions 
such as piecewise rigidity and weak-perspective projection. 
In practice, the source and target often differ substantially in local geometry 
(e.g., limb thickness, torso proportions, or surface details) 
that skeletal parameters alone cannot account for. 
The offset parameters $\{\mathbf{o}_k\}$, 
released in the later optimization stage, 
provide additional representational capacity 
to accommodate such local shape discrepancies, 
reducing the model mismatch between the rig-based representation 
and the actual source appearance observed in the video.

\section{Extended Results}
\label{sec:appendix_perscene}



\begin{figure*}[h]
    \centering
    \includegraphics[width=0.85\linewidth,trim=0 9.9cm 1.9cm 0,clip]{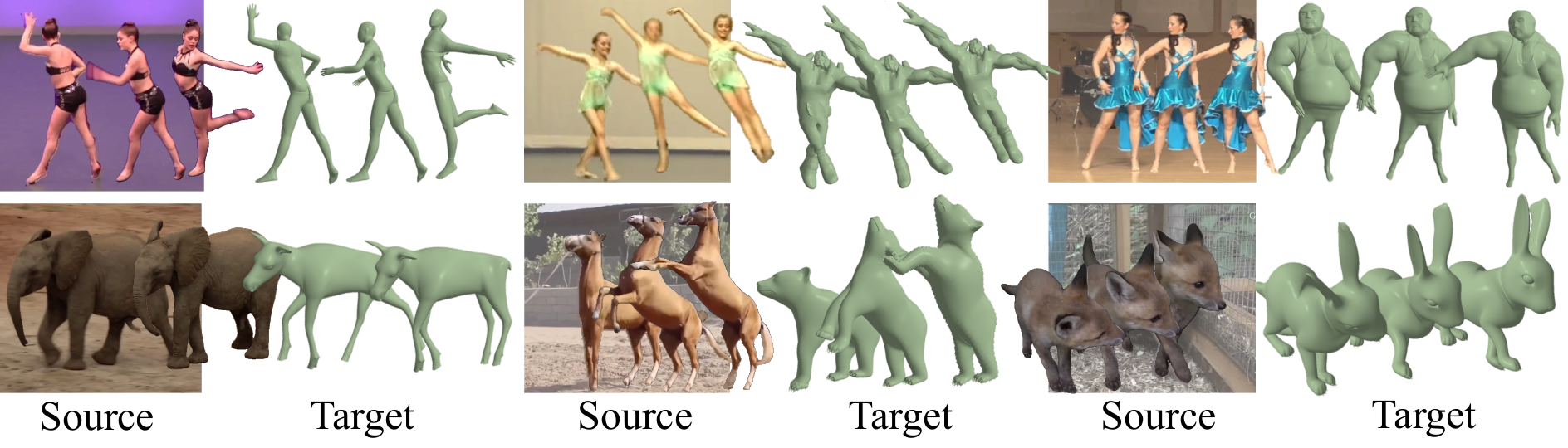}
    \caption{\textbf{Qualitative evaluation on extended real-world applicability.}}
    \label{fig:appendix_real}
\end{figure*}

\paragraph{\textbf{Extended real-world results.}} 
To further validate real-world applicability, we evaluate MorphGS on 8 additional human clips from SoloDance~\cite{solodance} and 8 animal clips from AiM~\cite{AiM}. For AiM scenes, we use the provided masks, whereas for SoloDance we extract foreground masks using SAM3~\cite{sam3} with the text query ``dancing person''.
As reported in Tab.~\ref{tab:quantitative-solodance} and Tab.~\ref{tab:quantitative-aim}, MorphGS consistently outperforms the baselines across both categories. Fig.~\ref{fig:appendix_real} further shows qualitatively that MorphGS transfers motion across source--target pairs with substantially different morphologies.

\begin{table*}[h]
  \centering
  \caption{\textbf{Quantitative evaluation on the SoloDance dataset.}  We use FVMD ($\times 10^{-3}$, $\downarrow$) as the quantitative metric for motion fidelity.}
  \vspace{-5pt}
  \label{tab:quantitative-solodance}
  {\scriptsize
  \renewcommand{\arraystretch}{1.05}
  \setlength{\tabcolsep}{3.2pt}
  \newcommand{\con}{\textcolor{gray!70}{--}}
  \begin{tabular}{@{}l cccccccc c@{}}
    \toprule
    Method & jazz\_1 & jazz\_2 & latin\_1 & latin\_2 & lyrical\_1 & lyrical\_2 & tap\_1 & tap\_2 & avg \\
    \midrule
    SPT$^+$       & 10.58 & 21.91 & 12.19 & 9.29 & 16.73 & 7.54 & 13.66 & 7.12 & 12.38 \\
    NPR$^+$       & 11.84 & 24.41 & 11.37 & 11.91 & 16.57 & 11.06 & 13.27 & 6.20 & 13.33 \\
    Pinocchio$^+$ & 9.80 & 24.27 & 12.73 & 9.95 & 17.15 & 7.70 & 14.31 & 6.57 & 12.81 \\
    \midrule
    Ours          & \textbf{9.12} & \textbf{18.15} & \textbf{9.26} & \textbf{5.05} & \textbf{14.63} & \textbf{7.53} & \textbf{12.38} & \textbf{5.90} & \textbf{10.25} \\
    \bottomrule
  \end{tabular}
  }
\end{table*}

\begin{table*}[h]
  \centering
  \caption{\textbf{Quantitative evaluation on the AiM dataset.}
  We use FVMD ($\times 10^{-3}$, $\downarrow$) as
  the quantitative metric for motion fidelity.}
  \label{tab:quantitative-aim}
  {\scriptsize
  \renewcommand{\arraystretch}{1.05}
  \setlength{\tabcolsep}{3.2pt}
  \newcommand{\con}{\textcolor{gray!70}{--}}
  \begin{tabular}{@{}l cccccccc c@{}}
    \toprule
    Method & boar & cow & hippo & horse & pig & elephant & tiger & rhino & avg \\
    \midrule
    NPR$^+$       & 10.72 & 27.00 & 13.71 & 18.01 & 9.06 & 15.76 & 26.80 & 7.77 & 16.10 \\
    Pinocchio$^+$ & 11.83 & 27.50 & 21.27 & 14.62 & 7.20 & 11.63 & 20.49 & 7.63 & 15.27 \\
    \midrule
    Ours          & \textbf{7.57} & \textbf{21.27} & \textbf{9.88} & \textbf{11.96} & \textbf{4.56} & \textbf{6.49} & \textbf{14.59} & \textbf{3.86} & \textbf{10.02} \\
    \bottomrule
  \end{tabular}
  }
\end{table*}

\begin{figure*}[t]
    \centering
    \includegraphics[width=0.85\linewidth,trim=0 0cm 2.2cm 0,clip]{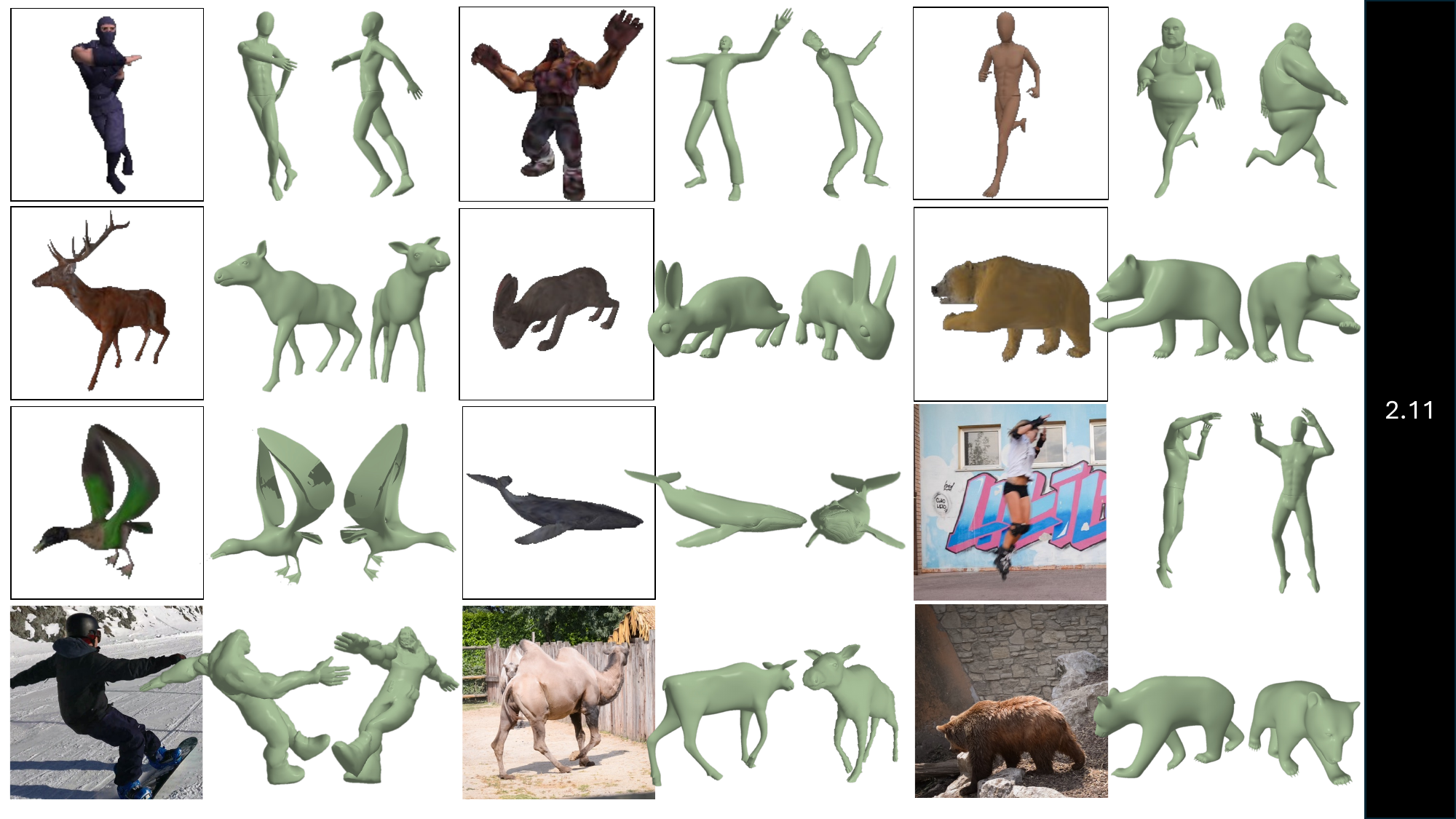}
    \caption{\textbf{3D consistency from off-axis views.}}
    \label{fig:appendix_sideview}
\end{figure*}

\paragraph{\textbf{3D coherence and per-scene quantitative results.}}
As MorphGS performs 2D-to-3D motion transfer, its output must hold up as a coherent 3D reconstruction beyond the input view. Fig.~\ref{fig:appendix_sideview} shows side-view renderings, where the recovered structure remains coherent under off-axis viewpoints for both synthetic and real-video examples.
We provide detailed quantitative results for all evaluation scenes from the Mixamo~\cite{mixamo} and DT4D~\cite{dt4d} datasets in Tab.~\ref{tab:appendix_mixamo}, Tab.~\ref{tab:appendix_dt4d}, and Tab.~\ref{tab:appendix_nonq}.

\begin{table*}[t]
    \vspace{-4pt}
    \caption{\footnotesize\textbf{Quantitative evaluation across all scenes from the Mixamo dataset.} Lower is better for both PMD and FVMD. Best and second-best results are highlighted in \cellbest{red} and \cellsecond{orange}, respectively.}
    \label{tab:appendix_mixamo}
    \setlength{\abovecaptionskip}{4pt}
    \setlength{\belowcaptionskip}{-2pt}
    \centering
    \resizebox{\textwidth}{!}{
    \begin{tabular}{l *{12}{c}}
    \toprule
    Method 
    & \multicolumn{2}{c}{JumpingJacks} 
    & \multicolumn{2}{c}{Running} 
    & \multicolumn{2}{c}{SideStep} 
    & \multicolumn{2}{c}{SkinningTest} 
    & \multicolumn{2}{c}{StandingJump} 
    & \multicolumn{2}{c}{SwingDance} \\
    & PMD & FVMD 
    & PMD & FVMD 
    & PMD & FVMD 
    & PMD & FVMD 
    & PMD & FVMD 
    & PMD & FVMD  \\
    \cmidrule(lr){1-1}
    \cmidrule(l){2-3}  \cmidrule(l){4-5}  \cmidrule(l){6-7}  \cmidrule(l){8-9}  \cmidrule(l){10-11}  \cmidrule(l){12-13}
    SPT$^+$           & \cellsecond{0.79} & \cellsecond{8.55} & \cellsecond{2.08} & 11.27 & 1.68 & \cellbest{6.68} & \cellsecond{2.48} & 20.14 & \cellbest{4.28} & 24.15 & 1.69 & \cellsecond{7.25} \\
    NPR$^+$           & 0.85 & 8.63 & 2.80 & 16.11 & 1.62 & 21.01 & 6.93 & 24.00 & 5.51 & 32.04 & 1.71 & 8.99 \\
    Transfer4D        & 6.14 & 16.27 & 5.66 & 17.92 & 1.61 & 25.50 & 5.56 & 18.77 & 9.62 & \cellsecond{24.12} & 2.56 & 11.09 \\
    Pinocchio$^+$       & 15.58 & 18.46 & 2.76 & \cellsecond{10.64} & \cellsecond{1.16} & {7.37} & 2.50 & \cellsecond{17.85} & \cellsecond{4.55} & 26.96 & \cellsecond{1.61} & 11.33 \\
    Ours              & \cellbest{0.39} & \cellbest{4.19} & \cellbest{1.54} & \cellbest{6.54} & \cellbest{1.03} & \cellsecond{6.84} & \cellbest{1.71} & \cellbest{17.66} & 5.08 & \cellbest{22.55} & \cellbest{1.04} & \cellbest{5.74} \\
    \bottomrule
    \toprule
    Method 
    & \multicolumn{2}{c}{Walking} 
    & \multicolumn{2}{c}{Floating} 
    & \multicolumn{2}{c}{HipHopDance} 
    & \multicolumn{2}{c}{Header} 
    & \multicolumn{2}{c}{Dying} 
    & \multicolumn{2}{c}{Snatch} \\
    & PMD & FVMD 
    & PMD & FVMD 
    & PMD & FVMD 
    & PMD & FVMD 
    & PMD & FVMD 
    & PMD & FVMD  \\
    \cmidrule(lr){1-1}
    \cmidrule(l){2-3}  \cmidrule(l){4-5}  \cmidrule(l){6-7}  \cmidrule(l){8-9}  \cmidrule(l){10-11}  \cmidrule(l){12-13}
    SPT$^+$           & \cellsecond{1.83} & \cellsecond{6.07} & 5.08 & 10.20 & 1.51 & 8.25 & \cellsecond{4.86} & {7.80} & 5.50 & 7.88 & \cellsecond{3.71} & \cellsecond{32.46} \\
    NPR$^+$           & 2.97 & 9.55 & 4.20 & 11.35 & 1.64 & 10.22 & 5.44 & 8.87 & 8.95 & 9.24 & 5.17 & 39.98 \\
    Transfer4D        & 7.16 & 12.44 & 13.89 & \cellbest{7.40} & 2.75 & 11.57 & 19.08 & \cellsecond{6.90} & 6.38 & 8.19 & 11.40 & 33.70 \\
    Pinocchio$^+$       & 2.32 & 6.83 & \cellsecond{2.79} & 11.10 & \cellsecond{1.36} & \cellsecond{7.80} & 5.20 & 11.32 & \cellsecond{3.15} & \cellbest{5.05} & 11.61 & 33.70 \\
    Ours              & \cellbest{0.93} & \cellbest{4.11} & \cellbest{1.97} & \cellsecond{7.63} & \cellbest{0.60} & \cellbest{6.82} & \cellbest{3.11} & \cellbest{6.81} & \cellbest{3.08} & \cellsecond{5.38} & \cellbest{2.44} & \cellbest{11.52} \\
    \bottomrule
    \end{tabular}
    }
\end{table*}


\begin{table*}[t]
{
\vspace{-4pt}
\caption{\footnotesize 
\textbf{Quantitative evaluation across all scenes from DT4D-Quadrupeds.}
PMD and FVMD values are reported in $10^{3}$ and $10^{-3}$ scales, respectively. 
Lower is better. Best and second-best results are highlighted in \cellbest{red} and \cellsecond{orange}, respectively.
}
\label{tab:appendix_dt4d}
    {\fontsize{8.5pt}{9.5pt}\selectfont
    \setlength{\tabcolsep}{2.8pt} 
    \setlength{\abovecaptionskip}{4pt}
    \setlength{\belowcaptionskip}{-2pt}
    \centering
    \begin{tabularx}{\textwidth}{l *{10}{Y}}
        \toprule
Method & \multicolumn{2}{c}{Punch} & \multicolumn{2}{c}{Walk1} & \multicolumn{2}{c}{Death} & \multicolumn{2}{c}{Walk2} & \multicolumn{2}{c}{KickBack} \\
       & PMD & FVMD & PMD & FVMD & PMD & FVMD & PMD & FVMD & PMD & FVMD \\
        \cmidrule(lr){1-1} \cmidrule(lr){2-3} \cmidrule(lr){4-5} \cmidrule(lr){6-7} \cmidrule(lr){8-9} \cmidrule(lr){10-11} 
NPR$^+$       & \cellsecond{2.92} & \cellsecond{8.48} & 2.32 & 23.23 & 5.31 & 27.51 & \cellsecond{1.27} & \cellsecond{8.64} & \cellsecond{2.95} & \cellsecond{20.61} \\
Transfer4D    & 4.54 & 10.40 & \cellsecond{1.96} & 21.45 & 5.31 & {23.21} & 8.74 & 12.06 & 3.56 & 20.64 \\
Pinocchio$^+$ & 5.50 & 9.11 & 3.20 & \cellsecond{19.75} & \cellsecond{5.20} & \cellsecond{20.40 }& 1.60 & 9.31 & 3.70 & {24.38} \\
Ours          & \cellbest{0.77} & \cellbest{6.81} & \cellbest{0.56} & \cellbest{8.62} & \cellbest{3.96} & \cellbest{14.12} & \cellbest{0.60} & \cellbest{4.92} & \cellbest{1.47} & \cellbest{13.70} \\
        \bottomrule
        \toprule
Method & \multicolumn{2}{c}{Swim} & \multicolumn{2}{c}{Jump} & \multicolumn{2}{c}{Walk3} & \multicolumn{2}{c}{Aggression} & \multicolumn{2}{c}{Howl}  \\
       & PMD & FVMD & PMD & FVMD & PMD & FVMD & PMD & FVMD & PMD & FVMD \\
        \cmidrule(lr){1-1} \cmidrule(lr){2-3} \cmidrule(lr){4-5} \cmidrule(lr){6-7} \cmidrule(lr){8-9} \cmidrule(lr){10-11} 
NPR$^+$       & \cellsecond{3.56} & 12.03 & \cellsecond{2.05} & \cellsecond{5.65} & 3.92 & 18.24 & \cellsecond{3.51} & {25.66} & \cellsecond{3.38} & 16.15 \\
Transfer4D    & 5.21 & 12.31 & 3.11 & 8.19 & \cellsecond{3.39} & \cellsecond{7.60} & 4.87 & 31.32 & 4.26 & \cellsecond{7.00} \\
Pinocchio$^+$ & 4.80 & \cellsecond{7.83} & 16.30 & 6.71 & 10.30 & 17.73 & 6.60 & \cellsecond{25.63} & 7.90 & 14.00 \\
Ours          & \cellbest{2.23} & \cellbest{7.61} & \cellbest{1.66} & \cellbest{2.71} & \cellbest{0.75} & \cellbest{4.94} & \cellbest{1.26} & \cellbest{18.72} & \cellbest{1.05} & \cellbest{6.11} \\
        \bottomrule
        \toprule
Method & \multicolumn{2}{c}{Hit Back} & \multicolumn{2}{c}{Run Stop} & \multicolumn{2}{c}{Run Forward} & \multicolumn{2}{c}{Drink} & \multicolumn{2}{c}{Hop Forward} \\
       & PMD & FVMD & PMD & FVMD & PMD & FVMD & PMD & FVMD & PMD & FVMD \\
        \cmidrule(lr){1-1} \cmidrule(lr){2-3} \cmidrule(lr){4-5} \cmidrule(lr){6-7} \cmidrule(lr){8-9} \cmidrule(lr){10-11} 
NPR$^+$       & \cellsecond{2.25} & \cellsecond{10.02} & 5.61 & 12.39 & 4.74 & {5.02} & \cellsecond{2.80} & 27.67 & {2.55} & 23.48 \\
Transfer4D    & 5.29 & 13.72 & 4.22 & 23.86 & 8.83 & 5.24 & 5.28 & \cellsecond{17.13} & \cellsecond{1.44} & \cellsecond{15.91} \\
Pinocchio$^+$ & 3.40 & 17.07 & \cellsecond{2.90} & \cellsecond{10.89} & \cellsecond{4.20} & \cellsecond{4.28} & 4.30 & 20.51 & 3.30 & 21.07 \\
Ours          & \cellbest{0.88} & \cellbest{9.32} & \cellbest{1.22} & \cellbest{8.36} & \cellbest{2.39} & \cellbest{2.92} & \cellbest{0.65} & \cellbest{16.45} & \cellbest{0.51} & \cellbest{6.80} \\
        \bottomrule
    \end{tabularx}
    }
}
\end{table*}


\begin{table*}[t]
{
\vspace{-4pt}
\caption{\footnotesize 
\textbf{Quantitative evaluation across all scenes from the DT4D-others dataset.}
Lower is better for both PMD ($\downarrow$) and FVMD ($\downarrow$). Best and second-best results are highlighted in \cellbest{red} and \cellsecond{orange}, respectively.}
\label{tab:appendix_nonq}
    {\fontsize{8.5pt}{9.5pt}\selectfont
    \setlength{\tabcolsep}{2.2pt}
    \setlength{\abovecaptionskip}{4pt}
    \setlength{\belowcaptionskip}{-2pt}
    \resizebox{\textwidth}{!}{
    \centering
    \begin{tabularx}{\textwidth}{Y *{5}{cc}}
        \toprule
    Method 
  & \multicolumn{2}{c}{Fly} 
  & \multicolumn{2}{c}{Attack} 
  & \multicolumn{2}{c}{Running} 
  & \multicolumn{2}{c}{Walk} 
  & \multicolumn{2}{c}{Swimming} \\
   & PMD & FVMD
   & PMD & FVMD
   & PMD & FVMD 
   & PMD & FVMD
   & PMD & FVMD  \\
    \cmidrule(lr){1-1}
    \cmidrule(l){2-3}  \cmidrule(l){4-5}  \cmidrule(l){6-7}  \cmidrule(l){8-9}  \cmidrule(l){10-11}
    Transfer4D    & 13.21 & 45.51 & 2.67 & 21.40 & \cellsecond{14.08} &
    \cellbest{11.15} & 3.59 & \cellsecond{3.80} & \cellsecond{2.25} & 5.74 \\
    Pinocchio$^+$  & \cellsecond{2.91} & \cellbest{30.63} & \cellsecond{2.14} & \cellsecond{11.01} & 29.99 & 17.37  & \cellbest{0.43} & 8.97  & 2.45 & \cellbest{3.18} \\
    Ours   & \cellbest{1.56} & \cellsecond{36.07} & \cellbest{1.81} & \cellbest{8.81} & \cellbest{5.19} & \cellsecond{11.26} &
    \cellsecond{0.57} & \cellbest{3.36} & \cellbest{0.62} & \cellsecond{4.63} \\
    \bottomrule
\end{tabularx}
    }
    }
}
\end{table*}

\end{document}